\documentclass{article}

\PassOptionsToPackage{numbers,compress}{natbib}
\usepackage[main,final]{neurips_2026}

\usepackage[utf8]{inputenc}
\usepackage[T1]{fontenc}
\usepackage{hyperref}
\usepackage{url}
\usepackage{booktabs}
\usepackage{amsfonts}
\usepackage{amsmath,amssymb}
\usepackage{nicefrac}
\usepackage{microtype}
\usepackage{xcolor}
\usepackage{graphicx}
\usepackage{subcaption}
\usepackage{multirow}
\usepackage{array}
\usepackage{enumitem}
\usepackage{siunitx}
\usepackage{placeins}
\usepackage{float}
\usepackage{colortbl}

\definecolor{semroad}{rgb}{0.30,0.30,0.35}
\definecolor{semwalk}{rgb}{0.45,0.45,0.40}
\definecolor{semcenter}{rgb}{0.90,0.80,0.10}
\definecolor{semstatic}{rgb}{0.80,0.40,0.10}
\definecolor{semvehicle}{rgb}{0.20,0.50,0.90}

\newcommand{\semlegenditem}[2]{%
    \raisebox{0.15ex}{%
        \colorbox{#1}{\rule{0pt}{0.85em}\hspace{1.2em}}%
    }%
    \hspace{0.35em}#2%
}

\title{Trajectory Evaluation via Rollout-Free World Model for End-to-end Autonomous Driving}

\author{
  Junzhe Wang$^{1}$\thanks{Equal contribution.} \quad
  Jinqing Zhang$^{1,2}$\footnotemark[1] \quad
  Qingjie Liu$^{1}$\thanks{Corresponding author.} \\
  $^{1}$State Key Laboratory of Virtual Reality Technology and Systems,\\
  Beihang University, Beijing, China\\
  $^{2}$Xiaomi EV\\
  \texttt{qingjie.liu@buaa.edu.cn}
}

\begin{document}

\maketitle

\begin{abstract}
Although existing candidate trajectory evaluators have substantially improved end-to-end planning accuracy, their underlying mechanism remains suboptimal. Model-based evaluators typically make scoring future-aware by explicitly predicting candidate-conditioned future states over the planning horizon, which we term rollout. However, rollout incurs high inference cost and accumulates prediction errors. We argue that planning does not require reconstructing a full future scene for each candidate, but only future-relevant information for reliable comparison. We therefore propose \textbf{RFWorld}, a Rollout-Free World model for end-to-end trajectory evaluation. RFWorld constructs a shared scene memory and shapes it into a future-readable representation via lightweight training-time temporal readout with semantic supervision, keeping training overhead low. At inference time, candidate trajectories directly query the shared memory via sparse spatial readout for scoring, reducing computation by roughly the planning horizon times the number of candidate trajectories, while avoiding recursive rollout errors. Experiments on NAVSIM and Bench2Drive show that RFWorld learns a more planning-aligned world representation and achieves state-of-the-art performance.
\end{abstract}

\section{Introduction}

\begin{figure}[t]
    \centering
    \includegraphics[width=\linewidth]{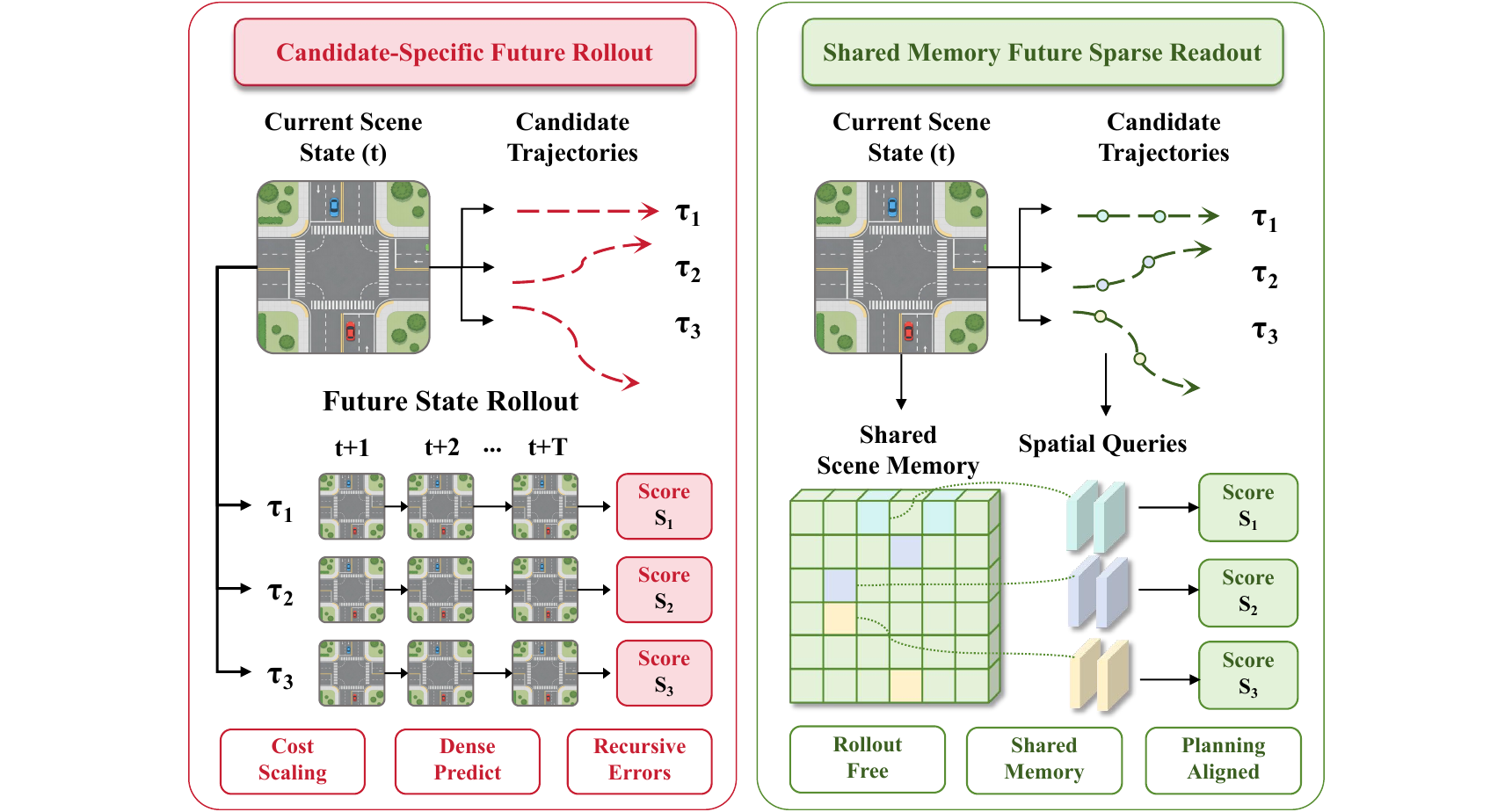}
    \caption{\textbf{Two paradigms for future-aware trajectory evaluation.}
    Conventional methods evaluate each candidate by rolling out its future states separately. In contrast, RFWorld constructs a shared scene memory once and lets all candidate trajectories read from it directly for rollout-free scoring.}
    \label{fig:paradigm}
\end{figure}

End-to-end autonomous driving~\cite{chen2024e2eadsurvey,chitta2023transfuser,sadat2020p3,hu2022stp3,hu2023uniad,jiang2023vad} has attracted increasing attention by directly mapping sensor observations to future trajectories or low-level control commands, reducing the interface bottlenecks and error accumulation of conventional modular pipelines. Recent planning-oriented methods~\cite{chen2024vadv2,sun2025sparsedrive,liao2025diffusiondrive,xing2025goalflow,zhang2025carplanner} have further advanced this paradigm from single-trajectory regression to anchor-based, proposal-based, diffusion-based, and flow-based trajectory generation. However, as the action space becomes increasingly multi-modal, a fundamental question remains: given multiple plausible future motions, how should an autonomous driving system evaluate them before execution?

Trajectory evaluation is indispensable for safe and reliable end-to-end planning. Classical methods~\cite{fan2018apollo,treiber2000congested,dauner2023pdmclosed} rely on hand-crafted rules, cost functions, or model predictive objectives. More recent learning-based approaches~\cite{li2024hydra,li2025wote,shang2025drivedpo,yao2026drivesuprim} incorporate rule-based teachers, simulator-derived feedback, or learned score/reward models to distinguish unsafe candidates from human-like but risky ones. Yet many existing evaluators either score trajectories from the current state alone or depend on explicit candidate-conditioned future prediction.

A natural way to make trajectory evaluation future-aware is to introduce world models. Driving world models~\cite{chen2021worldonrails,hu2023gaia1,wang2023drivedreamer,wang2024drivewm,gao2024vista,zheng2024occworld,zhang2024bevworld,chen2025drivinggpt,zeng2025fsdrive,zhang2026resworld} have been explored for future image generation, occupancy forecasting, BEV-state prediction, latent dynamics modeling, and planning-oriented reasoning. In trajectory evaluation, they can imagine the future consequences of each candidate trajectory and estimate its reward or safety score~\cite{henaff2019mppuu,qiao2024marlcce,li2025wote}. Existing model-based evaluators typically do so by explicitly predicting candidate-conditioned future states over the planning horizon, which we term \emph{rollout}. However, should future-aware evaluation require rolling out a future scene for every candidate?

We argue that explicit candidate-wise rollout is misaligned with the actual needs of planning. Its inference cost grows with both the planning horizon and the number of candidates, while recursive prediction also makes trajectory scoring vulnerable to accumulated errors. More importantly, planning does not require reconstructing a full future scene for each candidate, but only accessing future-relevant information for reliable comparison.

Motivated by this, we reformulate model-based trajectory evaluation from trajectory-specific future rollout to rollout-free shared-memory readout, as illustrated in Fig.~\ref{fig:paradigm}. Instead of generating candidate-conditioned future states at inference time, RFWorld constructs a shared scene memory. To make all candidates readable through a common representation, we express their waypoints on the current ego-coordinate BEV plane rather than in independent ego-centric futures, enabling aligned spatial access to a unified memory. During training, temporal readout with semantic supervision shapes this memory into a future-readable representation with predictive temporal structure; during inference, candidate trajectories directly query it through sparse spatial readout for scoring.

This reformulation makes the world model more planning-aligned. Since the shared memory is computed once and reused across candidates, RFWorld avoids candidate-wise dense future generation, substantially reduces inference computation, keeps training overhead low through lightweight temporal shaping, and is less sensitive to compounding rollout errors. In this sense, RFWorld shifts the role of the world model from candidate-conditioned future simulation to constructing a planning-oriented shared memory for trajectory evaluation.

On NAVSIM~\cite{dauner2024navsim}, RFWorld achieves state-of-the-art performance and validates the effectiveness of rollout-free shared-memory trajectory evaluation. We also evaluate RFWorld in closed-loop simulation on the CARLA-based Bench2Drive~\cite{jia2024bench} benchmark.

Our contributions are summarized as follows:
\begin{itemize}[leftmargin=1.5em]
    \item We identify a key mismatch in future-aware trajectory evaluation: existing model-based evaluators typically rely on explicit candidate-conditioned future rollout, whereas end-to-end planning only requires future-relevant information to be directly queryable through a planning-sufficient interface.
    \item We propose \textbf{RFWorld}, a Rollout-Free World model for end-to-end trajectory evaluation. RFWorld constructs a shared scene memory, shapes it through lightweight training-time temporal readout with semantic supervision, and performs inference-time trajectory scoring via sparse spatial readout, thereby substantially reducing inference cost, keeping training overhead low, and mitigating accumulated rollout errors.
    \item We conduct extensive experiments on the NAVSIM benchmark and show that RFWorld achieves state-of-the-art performance, demonstrating the effectiveness of rollout-free shared-memory world modeling for end-to-end autonomous driving.
\end{itemize}
\section{Related Work}

\subsection{End-to-end Autonomous Driving}
End-to-end autonomous driving directly maps sensor observations to future trajectories or control commands, avoiding the interface bottlenecks of traditional modular pipelines. Representative works such as TransFuser \cite{chitta2023transfuser}, UniAD \cite{hu2023uniad}, and VAD \cite{jiang2023vad} establish the importance of sensor fusion, planning-oriented task design, and vectorized scene representations. A broad line of research further improves BEV representations through semantic-aware or geometry-aware learning and auxiliary tasks such as online HD map construction, temporal BEV fusion, and occupancy prediction \cite{li2022hdmapnet,huang2022bevdet4d,sima2023sceneasoccupancy,yu2023flashocc,zhang2023sa,zhang2025geobev,yuan2024streammapnet}. Recent studies \cite{sun2025sparsedrive,li2025ssr,guo2025ipad,chen2024vadv2,liao2025diffusiondrive,xing2025goalflow,zhang2025carplanner} further improve planning with sparse scene tokens, proposal-centric representations, anchor vocabularies, diffusion policies, flow matching, and reinforcement learning. Most of these methods focus on trajectory generation from current observations, while future-aware candidate evaluation remains less explored.

\subsection{Trajectory Evaluation}
Trajectory evaluation is crucial for safety-critical autonomous driving, especially when multiple candidate trajectories are available. Rule-based planners and simulators assess candidates using metrics such as collision, drivable-area compliance, progress, time-to-collision, and comfort. Recent end-to-end methods incorporate these signals into learning-based frameworks. Hydra-MDP \cite{li2024hydra} distills human and rule-based teachers through multiple score heads, DriveDPO \cite{shang2025drivedpo} points out that per-anchor score regression decouples supervision from direct policy optimization, and DriveSuprim \cite{yao2026drivesuprim} improves selection-based planning through coarse-to-fine filtering and self-distillation. Different from these methods, our work focuses on the future-aware representation and computation used for candidate scoring.

\subsection{World Model for Autonomous Driving}
World models have been increasingly explored for autonomous driving to model future scene evolution and support anticipatory planning. Image-level driving world models predict future visual observations for scenario generation or planning \cite{wang2024drivewm,chen2025drivinggpt,zeng2025fsdrive}, while occupancy- or latent-space models learn future scene dynamics in more compact representations \cite{zheng2024occworld,li2025law,zheng2025world4drive}. WoTE \cite{li2025wote} predicts future BEV states for trajectory evaluation and demonstrates the benefit of model-based scoring in BEV space. However, future-aware decision making in these methods is usually coupled with explicit future generation, latent-state prediction, or candidate-conditioned simulation. In contrast, our method shifts the computation from trajectory-specific rollout to shared-memory readout, using future supervision only to shape the memory during training and sparse spatial readout for inference-time trajectory evaluation.
\section{Preliminary}

\subsection{Candidate-based End-to-end Planning}

Given the observation at timestamp $t$, denoted as $\mathcal{O}_t$, the planner first generates a set of candidate trajectories
\[
\mathcal{T}_t=\{\tau_i\}_{i=1}^{N},
\]
where $\tau_i=\{p_i^k\}_{k=1}^{T}$ is the $i$-th candidate trajectory over $T$ future timestamps. The final planning output is selected by
\[
\tau^*=\arg\max_{\tau_i\in\mathcal{T}_t}s_i,
\]
where $s_i$ is the trajectory-level score of candidate $\tau_i$.

\subsection{Shared Memory and Spatio-Temporal Readout}

To evaluate candidate trajectories under different future conditions, the planner requires a representation that is both shared across candidates and accessible at specific temporal and spatial locations. We formulate such a representation as a shared memory:
\[
\mathcal{M}_t=\Phi_\theta(\mathcal{O}_t)\in\mathbb{R}^{L\times d},
\]
where $\Phi_\theta$ denotes the memory encoder, $L$ is the number of memory tokens, and $d$ is the feature dimension.

We use $t$ for the current absolute timestamp and $k$ for a relative future-step index, corresponding to time $t+k\Delta t$, where $\Delta t$ is the interval between consecutive steps.

The shared memory is accessed through query-conditioned readout. For future step $k$, a temporal query $q_k^{\mathrm{time}}$ reads out the corresponding time-conditioned feature:
\[
\bar{z}_k=\mathcal{R}(q_k^{\mathrm{time}},\mathcal{M}_t).
\]
During training, these temporal readouts are supervised by future-timestamp targets to shape the memory with predictive temporal information. This supervision encourages $\mathcal{M}_t$ to encode future-aware scene context, but it does not define the inference pathway.

During inference, candidate evaluation relies on spatial readout. For the waypoint $p_i^k$, we construct a spatial query
\[
q_{i,k}^{\mathrm{space}}=\psi(p_i^k),
\quad
z_{i,k}=\mathcal{R}(q_{i,k}^{\mathrm{space}},\mathcal{M}_t),
\]
where $\psi(\cdot)$ is a positional encoding function and $\mathcal{R}(\cdot)$ denotes an abstract memory readout operator. The concrete interaction mechanism is introduced in the method section. In this way, all candidate trajectories access the same temporally-shaped memory through their waypoint-conditioned spatial queries, without constructing independent future representations for each candidate.

\subsection{Planning-Sufficient Interface}

Under this formulation, the downstream planner operates on a relatively independent planning interface composed of the shared memory and the current ego state. Once $\mathcal{M}_t$ is constructed from the observation, the planner no longer revisits raw perceptual inputs, but evaluates candidate trajectories through memory readout with explicit ego-state conditioning:
\[
P_\theta(\tau^* \mid \mathcal{O}_t,\mathcal{T}_t)
=
P_\theta(\tau^* \mid \mathcal{M}_t,\mathcal{E}_t,\mathcal{T}_t),
\quad
\mathcal{M}_t=\Phi_\theta(\mathcal{O}_t).
\]
Here $P_\theta$ denotes the conditional distribution over the selected trajectory induced by the planning model, and $\mathcal{E}_t$ is the current ego state, which is directly fed into the reward encoder. Accordingly, the downstream planner depends on the original input only through the pair ($\mathcal{M}_t$,$\mathcal{E}_t$), rather than revisiting raw perceptual observations.

\section{Method}

\begin{figure}[t]
    \centering
    \includegraphics[width=\linewidth]{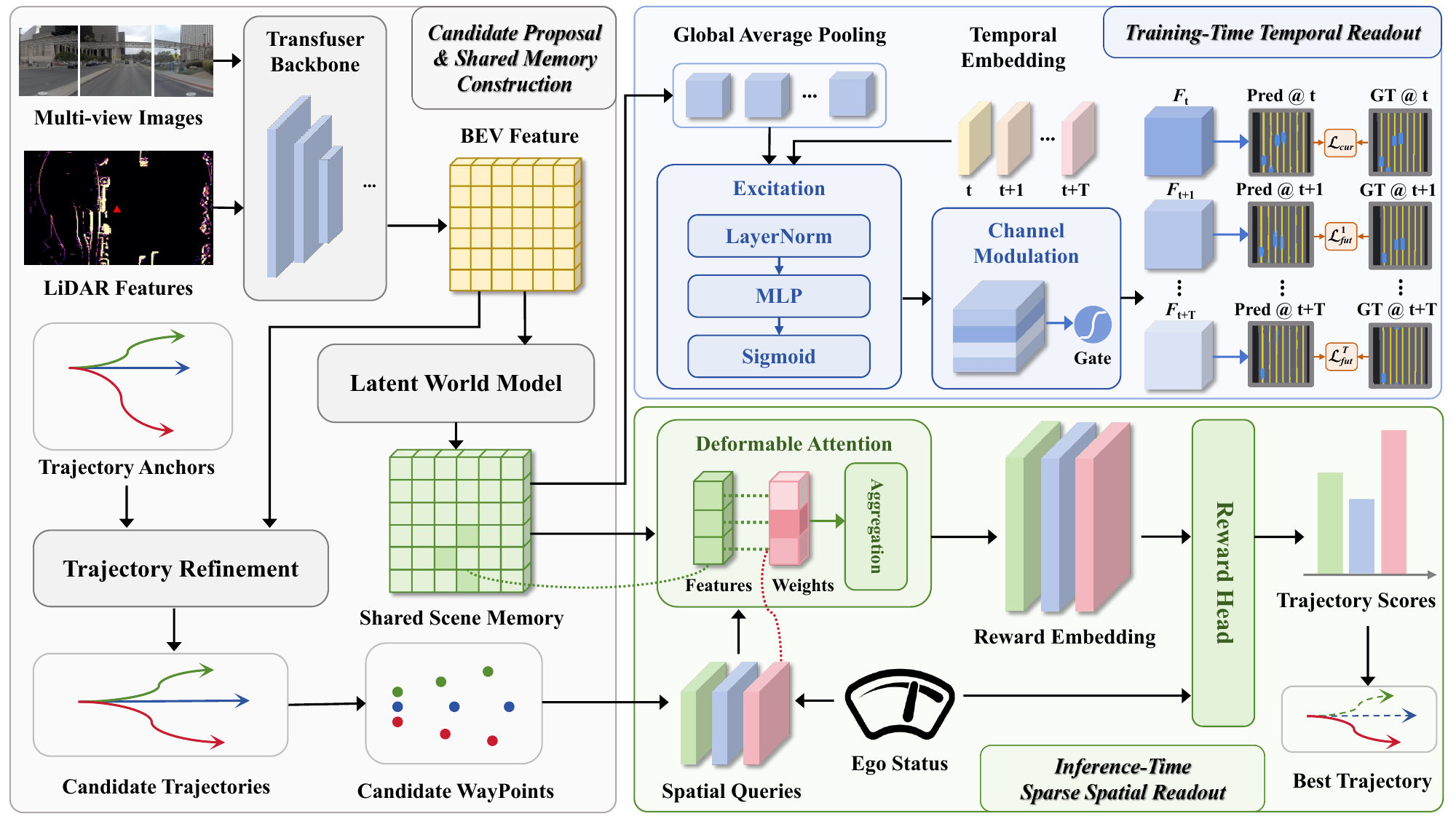}
    \caption{\textbf{Overall framework of RFWorld.} The current BEV representation is transformed into a shared memory. During training, temporal readout with semantic supervision shapes the memory into a future-aware representation. During inference, candidate trajectories directly query the shared memory for reward prediction, and the trajectory with highest score is selected as the final planning output.}
    \label{fig:overall_framework}
\end{figure}

Figure~\ref{fig:overall_framework} illustrates the overall architecture of RFWorld, which consists of candidate proposal, shared-memory construction with training-time temporal shaping, and inference-time trajectory evaluation via sparse spatial readout.

\subsection{Candidate Proposal and Shared Memory Construction}

Given the observation at timestamp $t$, we denote the model input as
\[
\mathcal{O}_t=\{\mathcal{H}_t,\mathcal{L}_t,\mathcal{E}_t,\mathcal{I}_t\},
\]
where $\mathcal{H}_t$ denotes historical frame information, $\mathcal{L}_t$ denotes the current LiDAR feature, $\mathcal{E}_t$ is the current ego state, and $\mathcal{I}_t=\{I_t^v\}_{v=1}^{N_c}$ denotes the current multi-view images. Following the multi-modal fusion paradigm of TransFuser~\cite{chitta2023transfuser}, we align historical LiDAR features to the current ego coordinate frame and fuse them with the current visual observations to obtain a BEV scene representation $B_t$. 

For planning, we adopt an anchor-based candidate proposal branch~\cite{chen2024vadv2,li2025wote}. Let $\mathcal{A}=\{a_i\}_{i=1}^{N}$ denote a predefined vocabulary of trajectory anchors. Each anchor interacts with the current scene representation to predict an offset, yielding the refined candidate trajectory
\[
\tau_i = a_i + \Delta_i, \quad i=1,\ldots,N.
\]

Different from candidate-conditioned world models that construct separate future representations for different trajectories~\cite{li2025wote,zhang2026resworld}, our method builds a trajectory-agnostic shared memory from the current BEV representation. Specifically, the current BEV feature $B_t$ is flattened, augmented with spatial positional embeddings, and processed by a Transformer-based~\cite{vaswani2017attention} latent world model to obtain
\[
\mathcal{M}_t=\Phi_{\mathrm{mem}}(\mathrm{Flat}(B_t)+P)\in\mathbb{R}^{L\times d},
\]
where $P$ denotes the spatial positional embedding, $L$ is the number of BEV tokens, and $d$ is the hidden dimension. The shared memory is conditioned only on the observation and is reused by both the training and inference branches.

\subsection{Training-Time Temporal Readout with Semantic Supervision}

To endow the shared memory with predictive temporal structure, we attach a lightweight temporal readout branch during training. Let
\[
\bar{\mathcal{M}}_t=\mathrm{Reshape}(\mathcal{M}_t)\in\mathbb{R}^{d\times H\times W}
\]
denote the spatial BEV form of the shared memory, where $H$ and $W$ are its height and width ($L=HW$), and let
\[
c_t=\mathrm{AvgPool}(\bar{\mathcal{M}}_t)\in\mathbb{R}^{d}
\]
be its globally pooled context. Let $\mathcal{S}$ be the selected supervised temporal indices, including $0$ for the current timestamp. For each $k\in\mathcal{S}$, we introduce a learnable temporal embedding $e_k$ and perform SE-style channel modulation~\cite{hu2018senet}:
\[
g_k = 2 \cdot \sigma\!\left(\mathrm{MLP}\!\left(\mathrm{LN}(c_t + e_k)\right)\right)\in\mathbb{R}^{d},
\]
\[
\bar{\mathcal{M}}_t^{(k)} = \bar{\mathcal{M}}_t \odot g_k.
\]
The gate $g_k$ is broadcast across spatial positions to reweight the memory channels. The resulting time-conditioned feature is decoded into a BEV semantic map by the shared semantic decoder $\mathcal{D}$:
\[
\hat{Y}_t^{(k)}=\mathcal{D}\!\left(\bar{\mathcal{M}}_t^{(k)}\right).
\]
The decoder is shared across all selected timestamps, so temporal diversity is induced solely by the temporal readout. The current prediction $\hat{Y}_t^{(0)}$ is supervised by the current semantic target, while the future predictions $\{\hat{Y}_t^{(k)}\}_{k\in\mathcal{S}\setminus\{0\}}$ are supervised by the selected future semantic targets. We denote the resulting semantic supervision as
\[
\mathcal{L}_{\mathrm{sem}}
=
\mathcal{L}_{\mathrm{cur}}
+
\sum_{k\in\mathcal{S}\setminus\{0\}} \omega_k\,\mathcal{L}_{\mathrm{fut}}^{(k)},
\]
where $\omega_k$ denotes the normalized weight of each supervised future timestamp. This branch is used only to shape the shared memory during training and does not define the inference-time planning path.

\subsection{Inference-Time Sparse Spatial Readout}

During inference, we do not explicitly decode future semantic maps or perform candidate-wise rollout. Instead, for each candidate trajectory $\tau_i=\{p_i^k\}_{k=1}^{T}$, we directly read the information required for trajectory evaluation from the shared memory through deformable spatial attention~\cite{zhu2021deformabledetr}. Specifically, each waypoint $p_i^k$ is encoded as a spatial query
\[
q_i^k=\psi(p_i^k),
\]
where $\psi(\cdot)$ denotes the waypoint encoder. Ego-state context is injected into the query features to modulate the readout. The offsets and attention weights are learned by $\Gamma$; bilinear interpolation in $V$ is only the differentiable operation used to sample at the learned continuous locations.

Given query $q_i^k$ and its reference point $r_i^k$ on the BEV plane, the reward encoder predicts a set of sampling offsets $\Delta_{i,k}^{m}$ and attention weights $\alpha_{i,k}^{m}$:
\[
\{\Delta_{i,k}^{m},\,\alpha_{i,k}^{m}\}_{m=1}^{M}
=\Gamma(q_i^k),
\]
where $\Gamma(\cdot)$ denotes the sampling-parameter prediction head, and $M$ is the number of sampling points. The waypoint feature is then obtained by sparse weighted sampling from $\bar{\mathcal{M}}_t$:
\[
z_i^k=\sum_{m=1}^{M}\alpha_{i,k}^{m}\,
V\!\left(\bar{\mathcal{M}}_t,\; r_i^k+\Delta_{i,k}^{m}\right),
\]
where $V(\cdot)$ denotes bilinear sampling on the shared memory. The waypoint features are aggregated along the trajectory horizon to produce a trajectory embedding $e_i$, which is further fused with ego-state context before reward prediction.

Following prior trajectory-evaluation frameworks~\cite{li2025wote}, we predict an imitation reward and a set of simulation rewards:
\[
r_i^{\mathrm{im}}=\mathrm{Head}_{\mathrm{im}}(e_i), \qquad
r_i^{\mathrm{sim}}=\mathrm{Head}_{\mathrm{sim}}(e_i).
\]
The former captures similarity to expert behavior, whereas the latter evaluates the candidate under simulator-derived safety and planning criteria. At inference time, these rewards are aggregated into a final trajectory score, and the highest-scoring trajectory is selected as the planning output.

\subsection{Training Objective}

We train the model with a compact multi-term objective consisting of trajectory, semantic, imitation-reward, and simulation-reward supervision. For trajectory prediction, only the candidate anchor closest to the ground-truth trajectory is supervised, yielding the trajectory loss $\mathcal{L}_{\mathrm{traj}}$. The semantic supervision in Sec.~4.2 is denoted as $\mathcal{L}_{\mathrm{sem}}$, where the semantic maps are supervised by a focal loss. For imitation reward learning, we construct a soft target distribution from the relative distances between candidate trajectories and the expert trajectory, and optimize the predicted imitation reward with a cross-entropy-style objective, yielding $\mathcal{L}_{\mathrm{im}}$. For simulation reward learning, the predicted heads for NC, DAC, EP, TTC, and Comfort are supervised by simulator-derived targets using BCE loss; the additional EP margin ranking regularization is absorbed into $\mathcal{L}_{\mathrm{sim}}$ for brevity.

The overall training objective is
\[
\mathcal{L}_{\mathrm{total}}
=
\lambda_{\mathrm{traj}} \mathcal{L}_{\mathrm{traj}}
+
\lambda_{\mathrm{sem}} \mathcal{L}_{\mathrm{sem}}
+
\lambda_{\mathrm{im}} \mathcal{L}_{\mathrm{im}}
+
\lambda_{\mathrm{sim}} \mathcal{L}_{\mathrm{sim}},
\]
where $\lambda_{\mathrm{traj}}$, $\lambda_{\mathrm{sem}}$, $\lambda_{\mathrm{im}}$, and $\lambda_{\mathrm{sim}}$ are the corresponding loss weights.

\section{Experiments}

\subsection{Benchmark}

\textbf{NAVSIM.}
We evaluate RFWorld on NAVSIM-v1~\cite{dauner2024navsim}, which is built upon real-world sensor data and emphasizes safety-critical and behaviorally diverse planning scenarios. The official trainval split contains 85,109 training samples and 18,179 validation samples; the official test split contains 12,146 scenarios. For each test scenario, the model selects exactly one final trajectory, and each metric is computed once for that scenario; reported values are averages over the 12,146 scenarios. The primary metric is the Predictive Driver Model Score (PDMS), which aggregates No At-Fault Collision (NC), Drivable Area Compliance (DAC), Ego Progress (EP), Time-to-Collision (TTC), and Comfort (Comf.). We report PDMS together with all constituent metrics.
For normalized subscores in $[0,1]$, NAVSIM-v1 computes $\mathrm{PDMS}=\mathrm{NC}\cdot\mathrm{DAC}\cdot(5\,\mathrm{TTC}+5\,\mathrm{EP}+2\,\mathrm{Comf.})/12$.

\textbf{Bench2Drive.}
We evaluate RFWorld on Bench2Drive~\cite{jia2024bench}, a closed-loop autonomous driving benchmark in CARLA 0.9.15, using the official 220 test routes. We report Success Rate (\%) and Driving Score.

\subsection{Implementation Details}

\textbf{NAVSIM.}
We train and evaluate RFWorld on NAVSIM-v1 under the official data split. For fair comparison with prior work, we adopt the same ResNet-34~\cite{he2016deep} backbone as Transfuser~\cite{chitta2023transfuser}. The model takes LiDAR features, ego states, and multi-view images as input, and predicts scores over 1024 candidate trajectory anchors. NAVSIM experiments are conducted on 2 RTX 3090 GPUs with a batch size of 16. We train the model for 50 epochs using AdamW with a base learning rate of $1\times10^{-4}$. Unless otherwise specified, these settings are shared across the NAVSIM experiments.

\textbf{Bench2Drive.}
The input consists of three front-facing cameras, four consecutive LiDAR frames, the ego state, and a navigation command. The three images are stitched and resized to $1024\times256$; LiDAR points are rasterized into a $256\times256$ BEV grid covering 32\,m in each direction at 0.25\,m resolution. We initialize from the original RFWorld pretrained weights and train with AdamW for 30,000 iterations using batch size 8, learning rate $5\times10^{-6}$, weight decay $10^{-4}$, and gradient clipping at 5. At inference, 1,024 candidate trajectories are ranked by their predicted rewards and the highest-scoring trajectory is selected at every simulation step. A vehicle PID controller converts it into steering, throttle, and brake commands by tracking the sixth trajectory point, with throttle capped at 0.55.

\subsection{Comparison with SOTA}

\textbf{NAVSIM.}
As shown in Table~\ref{tab:sota_navsim}, RFWorld achieves a PDMS of 90.5 under the ResNet-34 setting, with the highest PDMS, NC, and TTC among the compared learned methods. These results suggest that our future-aware modeling and trajectory evaluation pipeline can more accurately capture future scene evolution and more effectively suppress unsafe trajectory candidates.

\begin{table*}[t]
\centering
\small
\setlength{\tabcolsep}{6pt}
\caption{Comparison with state-of-the-art end-to-end driving methods on NAVSIM. The best results are denoted in \textbf{bold}. WM indicates whether the method explicitly employs a world model. For fair comparison, we report the scores using the same ResNet-34 backbone.}
\label{tab:sota_navsim}
\resizebox{\textwidth}{!}{%
\begin{tabular}{l|c|c|cccccc}
\toprule
Method & WM & Supervision & NC$\uparrow$ & DAC$\uparrow$ & TTC$\uparrow$ & Comf.$\uparrow$ & EP$\uparrow$ & PDMS$\uparrow$ \\
\midrule
Human & -- & -- & 100.0 & 100.0 & 99.9 & \textbf{100.0} & 87.5 & 94.8 \\
\midrule
Ego Status MLP & $\times$ & Human & 93.0 & 77.3 & 83.6 & \textbf{100.0} & 62.8 & 65.6 \\
VADv2~\cite{chen2024vadv2} & $\times$ & Human & 97.2 & 89.1 & 91.6 & \textbf{100.0} & 76.0 & 80.9 \\
UniAD~\cite{hu2023uniad} & $\times$ & Human & 97.8 & 91.9 & 92.9 & \textbf{100.0} & 78.8 & 83.4 \\
LTF~\cite{chitta2023transfuser} & $\times$ & Human & 97.4 & 92.8 & 92.4 & \textbf{100.0} & 79.0 & 83.8 \\
Transfuser~\cite{chitta2023transfuser} & $\times$ & Human & 97.7 & 92.8 & 92.8 & \textbf{100.0} & 79.2 & 84.0 \\
DRAMA~\cite{yuan2024drama} & $\times$ & Human & 98.0 & 93.1 & 94.8 & \textbf{100.0} & 80.1 & 85.5 \\
GoalFlow~\cite{xing2025goalflow} & $\times$ & Human & 98.3 & 93.8 & 94.3 & \textbf{100.0} & 79.8 & 85.7 \\
Hydra-MDP~\cite{li2024hydra} & $\times$ & Human \& Rule & 98.3 & 96.0 & 94.6 & \textbf{100.0} & 78.7 & 86.5 \\
Hydra-MDP++~\cite{li2025hydraplusplus} & $\times$ & Human \& Rule & 97.6 & 96.0 & 93.1 & \textbf{100.0} & 80.4 & 86.6 \\
DiffusionDrive~\cite{liao2025diffusiondrive} & $\times$ & Human & 98.2 & 96.2 & 94.7 & \textbf{100.0} & 82.2 & 88.1 \\
DriveSuprim~\cite{yao2026drivesuprim} & $\times$ & Human \& Rule & 97.8 & 97.3 & 93.6 & \textbf{100.0} & \textbf{86.7} & 89.9 \\
DriveDPO~\cite{shang2025drivedpo} & $\times$ & Human \& Rule & 98.5 & \textbf{98.1} & 94.8 & 99.9 & 84.3 & 90.0 \\
\midrule
LAW~\cite{li2025law} & $\checkmark$ & Human & 96.4 & 95.4 & 88.7 & 99.9 & 81.7 & 84.6 \\
WoTE~\cite{li2025wote} & $\checkmark$ & Human \& Rule & 98.5 & 96.8 & 94.9 & 99.9 & 81.9 & 88.3 \\
ResWorld~\cite{zhang2026resworld} & $\checkmark$ & Human & 98.9 & 96.5 & 95.6 & \textbf{100.0} & 83.1 & 89.0 \\
\rowcolor{gray!20}
RFWorld (Ours) & $\checkmark$ & Human \& Rule & \textbf{99.0} & 97.8 & \textbf{96.0} & \textbf{100.0} & 84.6 & \textbf{90.5} \\
\bottomrule
\end{tabular}%
}
\end{table*}

\textbf{Bench2Drive.}
As shown in Table~\ref{tab:bench2drive}, RFWorld achieves a Success Rate of 31.36\% and a Driving Score of 62.38, exceeding TCP by 1.36 percentage points and 2.48 score points, respectively, and leading the compared baselines on both metrics.

\begin{table}[t]
\centering
\small
\setlength{\tabcolsep}{8pt}
\renewcommand{\arraystretch}{1.08}
\caption{Closed-loop results on the official 220-route Bench2Drive~\cite{jia2024bench} test set.}
\label{tab:bench2drive}
\begin{tabular*}{0.6\textwidth}{@{\extracolsep{\fill}}lcc@{}}
\toprule
Method & Success Rate (\%)$\uparrow$ & Driving Score$\uparrow$ \\
\midrule
AD-MLP~\cite{zhai2023rethinking} & 0.00 & 18.05 \\
UniAD~\cite{hu2023uniad} & 16.36 & 45.81 \\
VAD~\cite{jiang2023vad} & 15.00 & 42.35 \\
TCP~\cite{wu2022tcp} & 30.00 & 59.90 \\
RFWorld & \textbf{31.36} & \textbf{62.38} \\
\bottomrule
\end{tabular*}
\end{table}

\subsection{Ablation Study}

\paragraph{Effect of world model and future supervision.}
We first study whether the improvement comes from introducing a world model itself or from learning a predictive representation with explicit future supervision. The four rows in Table~\ref{tab:ablation_wm_future_sup} differ only in the two indicated switches; the backbone, anchors, optimization schedule, split, and evaluation protocol are fixed. Directly adding the world model without future supervision does not improve performance, suggesting that a future branch optimized only through the final planning objective is insufficient to learn stable and informative predictive features.

In contrast, explicitly supervising future readout features consistently improves all major metrics. In the WM $\times$ / Future Sup. $\checkmark$ setting, temporal semantic readout supervises the backbone BEV feature directly, without the world model Transformer. Future supervision alone improves PDMS from 88.9 to 89.4, while the complete RFWorld model reaches 90.5, indicating complementary roles for future supervision and shared-memory world modeling.

\paragraph{Effect of the number of inference anchors.}
We study how increasing the number of inference-time candidate trajectories affects performance and cost. For fair comparison, both WoTE and RFWorld are trained with 1024 anchors and a batch size of 16, and different numbers of anchors are randomly sampled at test time. Due to their different training memory demands, WoTE is trained on two A6000 GPUs (48\,GB each), whereas RFWorld is trained on two RTX 3090 GPUs (24\,GB each). Inference latency is measured on a single RTX 3090 GPU.

As shown in Table~\ref{tab:ablation-anchor-infer}, RFWorld improves consistently as the candidate set grows from 256 to 1024, indicating that richer candidate coverage is more effectively utilized in RFWorld. This trend is consistent with the underlying complexity difference.  Let \(N\) denote the number of candidate trajectories, \(T\) the planning horizon, \(L\) the number of BEV tokens, and \(M\) the number of sampling points in sparse readout. WoTE scales as \(O\!\left(NT \cdot \mathcal{C}_{\mathrm{WM}}(L)\right)\), approximately \(O(NTL^{2}d)\) when transformer world-model computation dominates, because future prediction is repeated for every candidate over the whole horizon. In contrast, RFWorld constructs the shared memory once with cost \(O(\mathcal{C}_{\mathrm{WM}}(L))\) and scores candidates with sparse spatial readout at \(O(NTMd)\). Since \(M \ll L\), RFWorld removes the dominant \(NT\) factor from future-state generation. The same advantage appears in training memory: under the 1024-anchor, batch-size-16 setting, RFWorld uses 8.7\,GB GPU memory, compared with 33.9\,GB for WoTE. Consistently, RFWorld remains faster at all tested anchor counts, with latency increasing only from 35.1\,ms to 37.5\,ms as \(N\) grows from 256 to 1024, while keeping the evaluation protocol fixed across all settings.

\begin{table}[t]
\centering
\small
\setlength{\tabcolsep}{4.5pt}
\caption{Ablation study on world modeling and future supervision. ``WM'' denotes whether the world model branch is enabled, and ``Future Sup.'' denotes whether future readout features are explicitly supervised during training. The best results are shown in \textbf{bold}.}
\label{tab:ablation_wm_future_sup}
\begin{tabular}{c c c c c c c c}
\toprule
WM & Future Sup. & NC$\uparrow$ & DAC$\uparrow$ & TTC$\uparrow$ & Comf.$\uparrow$ & EP$\uparrow$ & PDMS$\uparrow$ \\
\midrule
$\times$ & $\times$ & 98.7 & 94.4 & 95.2 & 99.8 & 83.4 & 88.9 \\
$\times$ & $\checkmark$ & 98.8 & 97.1 & 95.4 & \textbf{100.0} & 83.7 & 89.4 \\
$\checkmark$ & $\times$ & 98.5 & 96.5 & 95.1 & 99.8 & 82.3 & 88.4 \\
$\checkmark$ & $\checkmark$ & \textbf{99.0} & \textbf{97.8} & \textbf{96.0} & \textbf{100.0} & \textbf{84.6} & \textbf{90.5} \\
\bottomrule
\end{tabular}
\end{table}

\begin{table}[t]
\centering
\small
\setlength{\tabcolsep}{4.2pt}
\caption{Ablation study on inference-time candidate anchors. ``\#Traj.'' denotes the number of candidate trajectories randomly sampled at test time. Latency is measured per sample on a single RTX 3090 GPU. The best results are shown in \textbf{bold}.}
\label{tab:ablation-anchor-infer}
\begin{tabular}{l c c c c c c c c}
\toprule
Method & \#Traj. & NC$\uparrow$ & DAC$\uparrow$ & TTC$\uparrow$ & Comf.$\uparrow$ & EP$\uparrow$ & PDMS$\uparrow$ & Latency$\downarrow$ \\
\midrule
WoTE~\cite{li2025wote}     & 256  & 98.4 & 96.7 & 94.6 & 99.9 & 82.3 & 88.3 & 37.0 ms \\
WoTE~\cite{li2025wote}     & 512  & 98.3 & 96.3 & 94.5 & 99.8 & 83.4 & 88.4 & 41.2 ms \\
WoTE~\cite{li2025wote}     & 1024  & 98.4 & 97.1 & 94.8 & \textbf{100.0} & 82.5 & 88.7 & 49.2 ms \\
\midrule
RFWorld  & 256  & 98.8 & 97.0 & 95.7 & 99.9 & 83.9 & 89.6 & 35.1 ms \\
RFWorld  & 512  & 98.9 & 97.7 & 95.5 & \textbf{100.0} & 84.4 & 90.2 & 36.2 ms \\
RFWorld  & 1024 & \textbf{99.0} & \textbf{97.8} & \textbf{96.0} & \textbf{100.0} & \textbf{84.6} & \textbf{90.5} & 37.5 ms \\
\bottomrule
\end{tabular}
\end{table}

\subsection{Qualitative Comparison}

\begin{figure*}[t]
    \centering
    \setlength{\tabcolsep}{2pt}

    \begin{subfigure}[t]{0.49\textwidth}
        \centering
        \makebox[\linewidth]{%
            \makebox[0.55\linewidth][c]{\footnotesize WoTE}%
            \makebox[0.5\linewidth][c]{\footnotesize RFWorld}%
        }
        \vspace{2pt}
        \includegraphics[width=\linewidth]{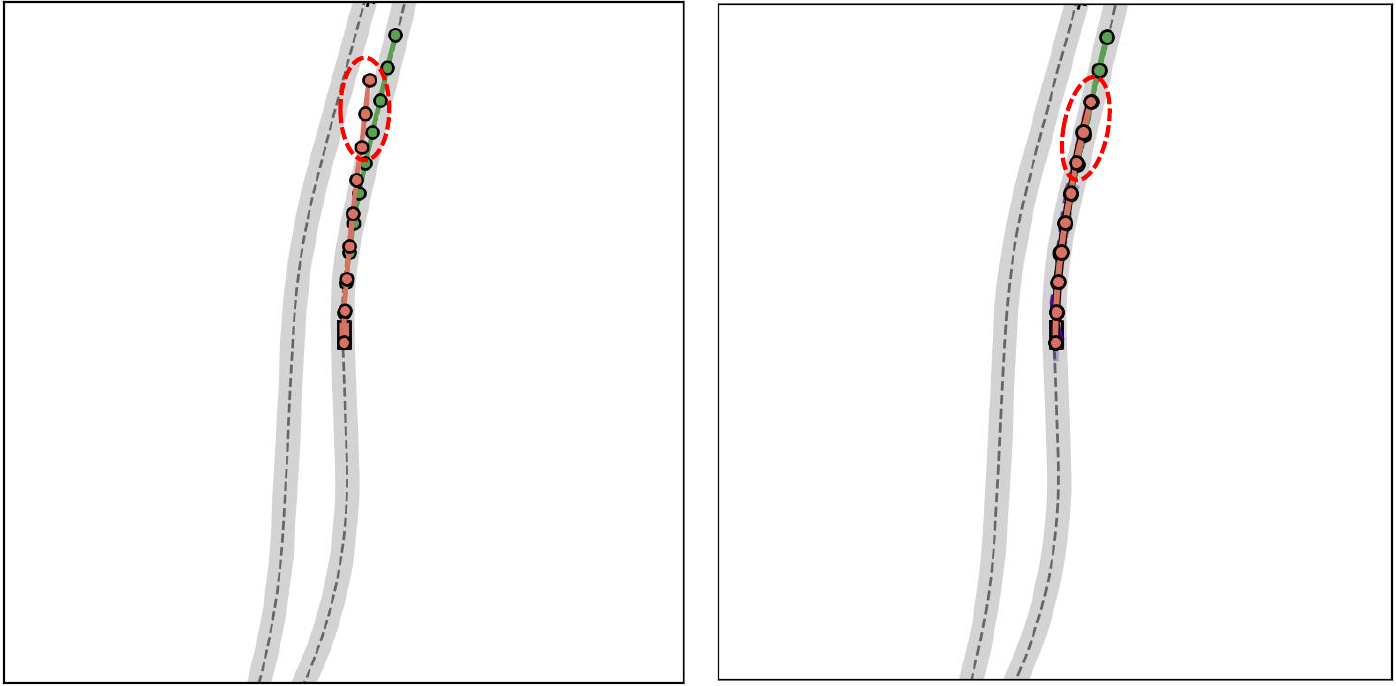}
        \caption{Avoiding drivable-area violation}
        \label{fig:qualitative_dac}
    \end{subfigure}
    \hfill
    \begin{subfigure}[t]{0.49\textwidth}
        \centering
        \makebox[\linewidth]{%
            \makebox[0.55\linewidth][c]{\footnotesize WoTE}%
            \makebox[0.5\linewidth][c]{\footnotesize RFWorld}%
        }
        \vspace{2pt}
        \includegraphics[width=\linewidth]{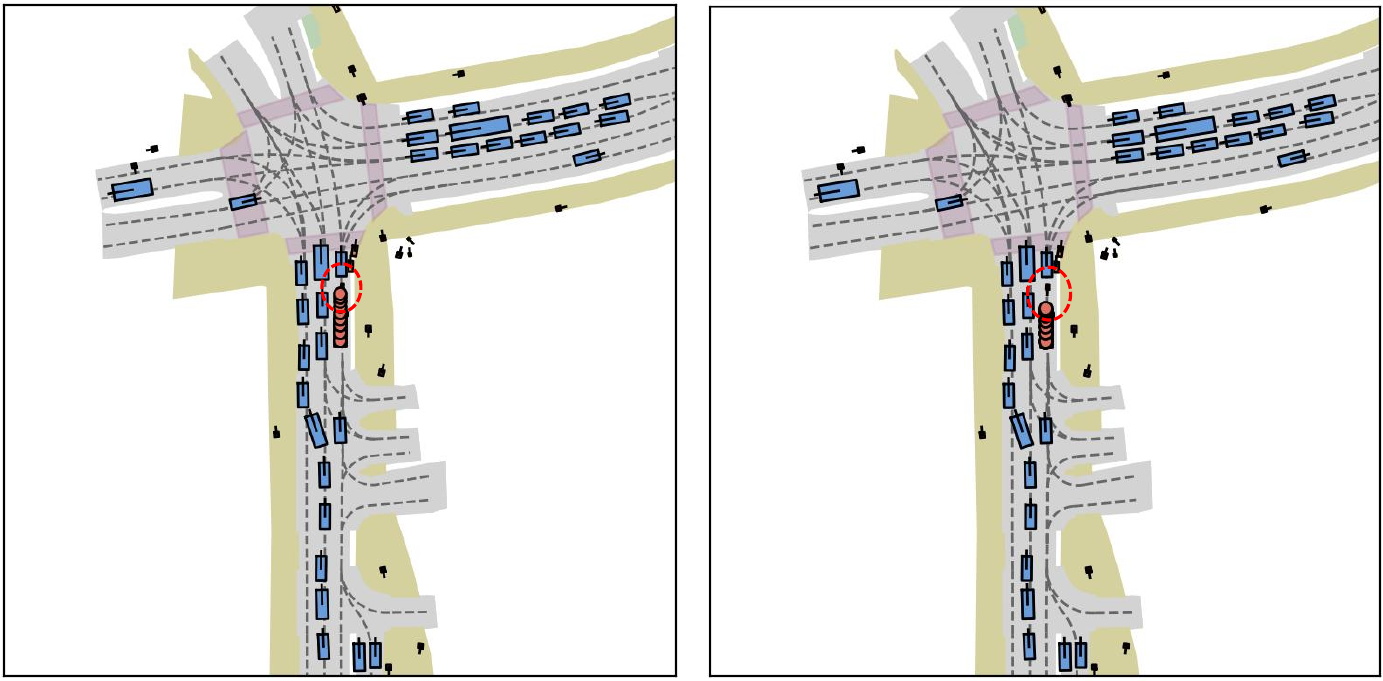}
        \caption{Avoiding collision with a pedestrian}
        \label{fig:qualitative_nc}
    \end{subfigure}

    \vspace{4pt}

    \begin{subfigure}[t]{0.49\textwidth}
        \centering
        \makebox[\linewidth]{%
            \makebox[0.55\linewidth][c]{\footnotesize WoTE}%
            \makebox[0.5\linewidth][c]{\footnotesize RFWorld}%
        }
        \vspace{2pt}
        \includegraphics[width=\linewidth]{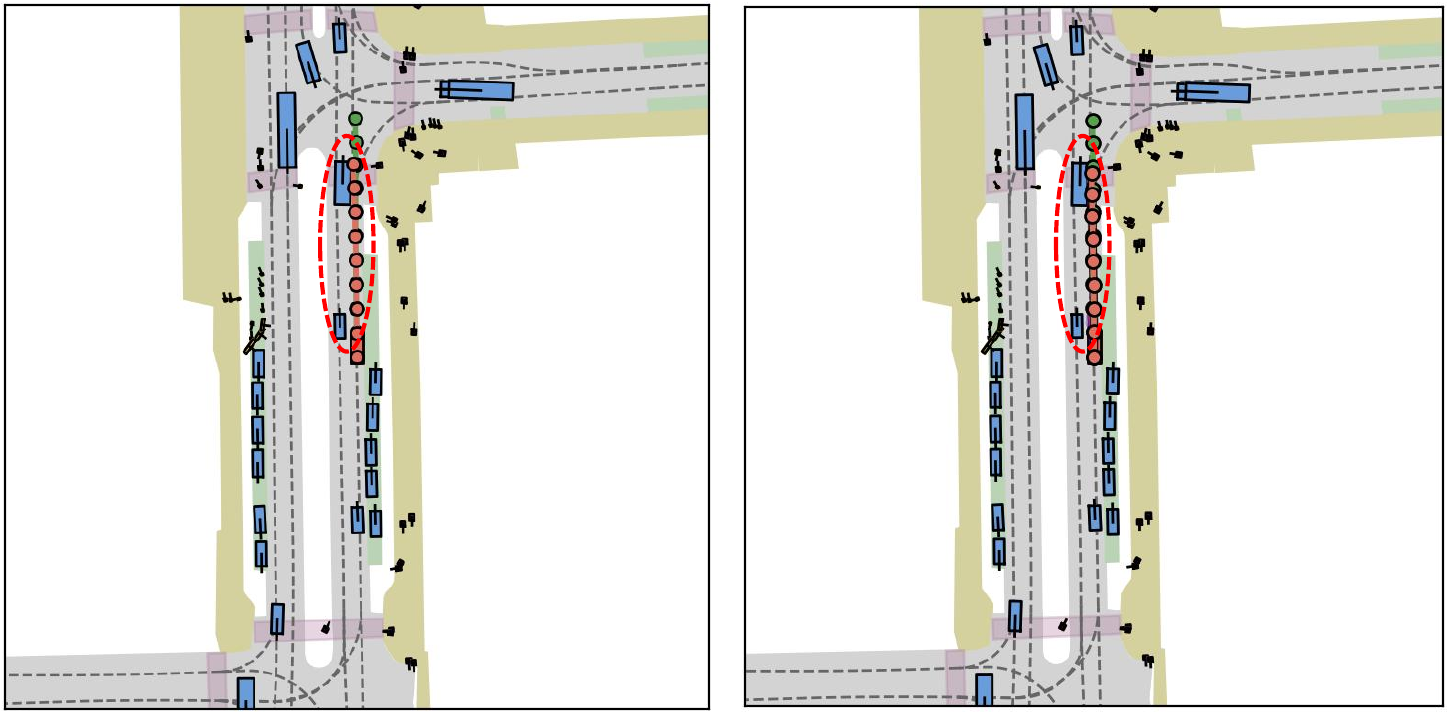}
        \caption{Avoiding side collision}
        \label{fig:qualitative_nc1}
    \end{subfigure}
    \hfill
    \begin{subfigure}[t]{0.49\textwidth}
        \centering
        \makebox[\linewidth]{%
            \makebox[0.55\linewidth][c]{\footnotesize WoTE}%
            \makebox[0.5\linewidth][c]{\footnotesize RFWorld}%
        }
        \vspace{2pt}
        \includegraphics[width=\linewidth]{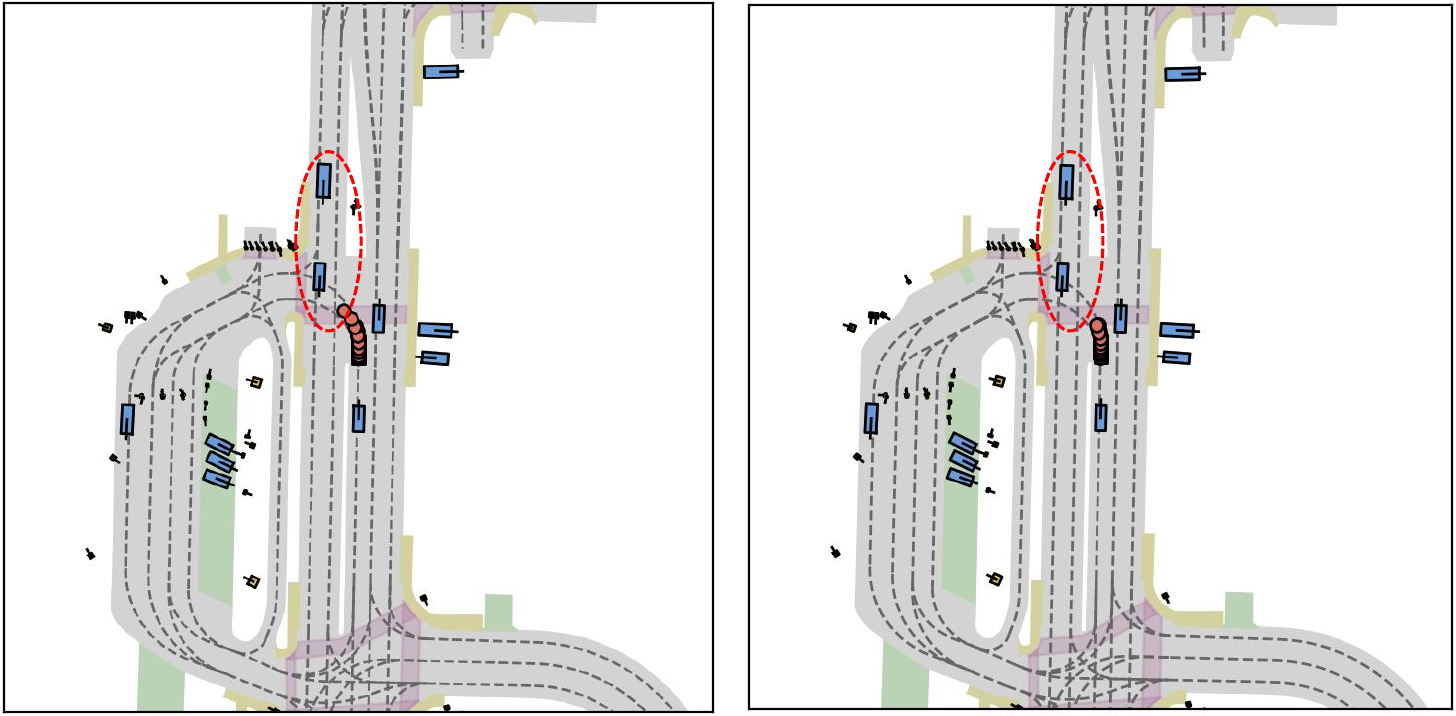}
        \caption{Yielding to oncoming traffic}
        \label{fig:qualitative_ttc}
    \end{subfigure}

    \vspace{2pt}
    \includegraphics[width=\textwidth]{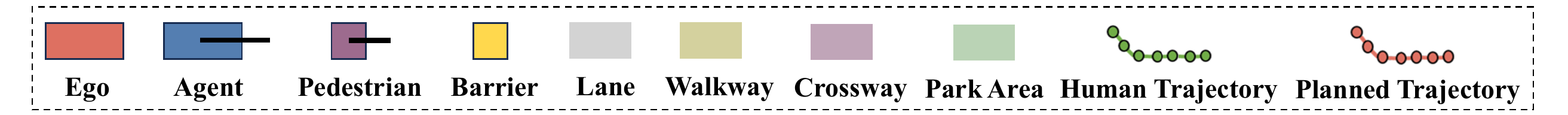}

    \caption{
        Qualitative comparison between WoTE~\cite{li2025wote} and RFWorld on NAVSIM-v1. Legend from DriveDPO~\cite{shang2025drivedpo}.
        (a) RFWorld avoids leaving the drivable area and remains within the lane boundaries.
        (b) RFWorld decelerates in advance and safely avoids the pedestrian.
        (c) RFWorld stays close to the lane centerline and avoids potential collision with the adjacent-lane vehicle.
        (d) RFWorld better accounts for oncoming traffic and yields before turning, thereby avoiding potential future risk.
    }
    \label{fig:qualitative_comparison}
\end{figure*}

We further provide qualitative comparisons with WoTE~\cite{li2025wote} in Fig.~\ref{fig:qualitative_comparison}. These cases indicate a broader advantage of RFWorld over rollout-based trajectory evaluation paradigms, where early rollout inaccuracies may accumulate over time and induce unreliable future assessments. In contrast, RFWorld better suppresses such compounding errors, leading to safer and more robust planning behaviors.

\section{Limitations and Future Work}

RFWorld has several limitations. First, its Transformer-based latent world model is not well suited to higher-resolution future prediction on dense BEV features; replacing dense BEV features with sparse scene tokens may improve both scalability and efficiency. Second, the quality of the shared memory remains dependent on the form of memory-shaping supervision, and the semantic targets are currently needed during training, motivating future work on pseudo-labeling and self-supervised temporal objectives. Third, observed failures mainly occur when the lightweight perception backbone produces inaccurate scene features or when relevant road geometry and traffic participants lie near or beyond the observation range. These errors can propagate to trajectory scoring and lead to drivable-area departures or collisions.

\section{Conclusion}

In this paper, we revisit model-based trajectory evaluation for end-to-end autonomous driving and identify a key mismatch in existing future-aware evaluators: they often rely on explicit candidate-conditioned future rollout, although trajectory selection only requires future-relevant information for candidate comparison. To address this issue, we propose RFWorld, a rollout-free world model for end-to-end trajectory evaluation. RFWorld constructs a shared scene memory, shapes it through training-time temporal readout with semantic supervision, and performs inference-time trajectory scoring via deformable spatial reward readout under explicit ego-state conditioning. Extensive experiments on the NAVSIM benchmark demonstrate that RFWorld provides a more planning-aligned world representation for future-aware trajectory evaluation and achieves state-of-the-art performance. These results show that replacing candidate-specific future rollout with shared-memory readout is an effective direction for end-to-end autonomous driving.

\begin{ack}
This research was supported by Zhejiang Provincial Natural Science Foundation of China under Grant No. LD24F020016 and National Natural Science Foundation of China under Grant No. 62576023.
The authors declare no competing interests.
\end{ack}

\bibliographystyle{plainnat}
\bibliography{references}

@article{chitta2023transfuser,
  title={{TransFuser}: Imitation with Transformer-Based Sensor Fusion for Autonomous Driving},
  author={Chitta, Kashyap and Prakash, Aditya and Jaeger, Bernhard and Yu, Zehao and Renz, Katrin and Geiger, Andreas},
  journal={IEEE Transactions on Pattern Analysis and Machine Intelligence},
  year={2023}
}

@inproceedings{hu2023uniad,
  title={Planning-Oriented Autonomous Driving},
  author={Hu, Yihan and Yang, Jiazhi and Chen, Li and Li, Keyu and Sima, Chonghao and Zhu, Xizhou and Chai, Siqi and Du, Senyao and Lin, Tianwei and Wang, Wenhai and Lu, Lewei and Jia, Xiaosong and Liu, Qiang and Dai, Jifeng and Qiao, Yu and Li, Hongyang},
  booktitle={Proceedings of the IEEE/CVF Conference on Computer Vision and Pattern Recognition},
  pages={17853--17862},
  year={2023}
}

@inproceedings{jiang2023vad,
  title={VAD: Vectorized Scene Representation for Efficient Autonomous Driving},
  author={Jiang, Bo and Chen, Shaoyu and Xu, Qing and Liao, Bencheng and Chen, Jiajie and Zhou, Helong and Zhang, Qian and Liu, Wenyu and Huang, Chang and Wang, Xinggang},
  booktitle={Proceedings of the IEEE/CVF International Conference on Computer Vision},
  pages={8340--8350},
  year={2023}
}

@article{chen2024vadv2,
  title={{VADv2}: End-to-End Vectorized Autonomous Driving via Probabilistic Planning},
  author={Chen, Shaoyu and Jiang, Bo and Gao, Hao and Liao, Bencheng and Xu, Qing and Zhang, Qian and Huang, Chang and Liu, Wenyu and Wang, Xinggang},
  journal={arXiv preprint arXiv:2402.13243},
  year={2024}
}

@inproceedings{sun2025sparsedrive,
  title={{SparseDrive}: End-to-End Autonomous Driving via Sparse Scene Representation},
  author={Sun, Wenchao and Lin, Xuewu and Shi, Yining and Zhang, Chuang and Wu, Haoran and Zheng, Sifa},
  booktitle={Proceedings of the IEEE International Conference on Robotics and Automation},
  year={2025}
}

@inproceedings{li2025ssr,
  title={Navigation-Guided Sparse Scene Representation for End-to-End Autonomous Driving},
  author={Li, Peidong and Cui, Dixiao},
  booktitle={International Conference on Learning Representations},
  year={2025}
}

@article{guo2025ipad,
  title={{iPad}: Iterative Proposal-centric End-to-End Autonomous Driving},
  author={Guo, Ke and Liu, Haochen and Wu, Xiaojun and Pan, Jia and Lv, Chen},
  journal={arXiv preprint arXiv:2505.15111},
  year={2025}
}

@inproceedings{liao2025diffusiondrive,
  title={DiffusionDrive: Truncated Diffusion Model for End-to-End Autonomous Driving},
  author={Liao, Bencheng and Chen, Shaoyu and Yin, Haoran and Jiang, Bo and Wang, Cheng and Yan, Sixu and Zhang, Xinbang and Li, Xiangyu and Zhang, Ying and Zhang, Qian and Wang, Xinggang},
  booktitle={Proceedings of the IEEE/CVF Conference on Computer Vision and Pattern Recognition},
  pages={12037--12047},
  year={2025}
}

@inproceedings{xing2025goalflow,
  title={{GoalFlow}: Goal-Driven Flow Matching for Multimodal Trajectories Generation in End-to-End Autonomous Driving},
  author={Xing, Zebin and Zhang, Xingyu and Hu, Yang and Jiang, Bo and He, Tong and Zhang, Qian and Long, Xiaoxiao and Yin, Wei},
  booktitle={Proceedings of the IEEE/CVF Conference on Computer Vision and Pattern Recognition},
  pages={1602--1611},
  year={2025}
}

@inproceedings{zhang2025carplanner,
  title={{CarPlanner}: Consistent Auto-regressive Trajectory Planning for Large-Scale Reinforcement Learning in Autonomous Driving},
  author={Zhang, Dongkun and Liang, Jiaming and Guo, Ke and Lu, Sha and Wang, Qi and Xiong, Rong and Miao, Zhenwei and Wang, Yue},
  booktitle={Proceedings of the IEEE/CVF Conference on Computer Vision and Pattern Recognition},
  pages={17239--17248},
  year={2025}
}

@article{li2024hydra,
  title={{Hydra-MDP}: End-to-End Multimodal Planning with Multi-Target Hydra-Distillation},
  author={Li, Zhenxin and Li, Kailin and Wang, Shihao and Lan, Shiyi and Yu, Zhiding and Ji, Yishen and Li, Zhiqi and Zhu, Ziyue and Kautz, Jan and Wu, Zuxuan and Jiang, Yu-Gang and Alvarez, Jose M.},
  journal={arXiv preprint arXiv:2406.06978},
  year={2024}
}

@inproceedings{shang2025drivedpo,
  title={{DriveDPO}: Policy Learning via Safety DPO for End-to-End Autonomous Driving},
  author={Shang, Shuyao and Chen, Yuntao and Wang, Yuqi and Li, Yingyan and Zhang, Zhaoxiang},
  booktitle={Advances in Neural Information Processing Systems},
  year={2025}
}

@inproceedings{yao2026drivesuprim,
  title={{DriveSuprim}: Towards Precise Trajectory Selection for End-to-End Planning},
  author={Yao, Wenhao and Li, Zhenxin and Lan, Shiyi and Wang, Zi and Sun, Xinglong and Alvarez, Jose M. and Wu, Zuxuan},
  booktitle={Proceedings of the AAAI Conference on Artificial Intelligence},
  volume={40},
  number={14},
  pages={11910--11918},
  year={2026}
}

@inproceedings{wang2024drivewm,
  title={Driving into the Future: Multiview Visual Forecasting and Planning with World Model for Autonomous Driving},
  author={Wang, Yuqi and He, Jiawei and Fan, Lue and Li, Hongxin and Chen, Yuntao and Zhang, Zhaoxiang},
  booktitle={Proceedings of the IEEE/CVF Conference on Computer Vision and Pattern Recognition},
  pages={14749--14759},
  year={2024}
}

@inproceedings{chen2025drivinggpt,
  title={{DrivingGPT}: Unifying Driving World Modeling and Planning with Multi-modal Autoregressive Transformers},
  author={Chen, Yuntao and Wang, Yuqi and Zhang, Zhaoxiang},
  booktitle={Proceedings of the IEEE/CVF International Conference on Computer Vision},
  pages={26890--26900},
  year={2025}
}

@inproceedings{zeng2025fsdrive,
  title={{FutureSightDrive}: Thinking Visually with Spatio-Temporal {CoT} for Autonomous Driving},
  author={Zeng, Shuang and Chang, Xinyuan and Xie, Mengwei and Liu, Xinran and Bai, Yifan and Pan, Zheng and Xu, Mu and Wei, Xing and Guo, Ning},
  booktitle={Advances in Neural Information Processing Systems},
  year={2025}
}

@inproceedings{zheng2024occworld,
  title={{OccWorld}: Learning a {3D} Occupancy World Model for Autonomous Driving},
  author={Zheng, Wenzhao and Chen, Weiliang and Huang, Yuanhui and Zhang, Borui and Duan, Yueqi and Lu, Jiwen},
  booktitle={European Conference on Computer Vision},
  year={2024}
}

@inproceedings{li2025law,
  title={Enhancing End-to-End Autonomous Driving with Latent World Model},
  author={Li, Yingyan and Fan, Lue and He, Jiawei and Wang, Yuqi and Chen, Yuntao and Zhang, Zhaoxiang and Tan, Tieniu},
  booktitle={International Conference on Learning Representations},
  year={2025}
}

@inproceedings{zheng2025world4drive,
  title={{World4Drive}: End-to-End Autonomous Driving via Intention-Aware Physical Latent World Model},
  author={Zheng, Yupeng and Yang, Pengxuan and Xing, Zebin and Zhang, Qichao and Zheng, Yuhang and Gao, Yinfeng and Li, Pengfei and Zhang, Teng and Xia, Zhongpu and Jia, Peng and Lang, XianPeng and Zhao, Dongbin},
  booktitle={Proceedings of the IEEE/CVF International Conference on Computer Vision},
  pages={28632--28642},
  year={2025}
}

@inproceedings{li2025wote,
  title={End-to-End Driving with Online Trajectory Evaluation via {BEV} World Model},
  author={Li, Yingyan and Wang, Yuqi and Liu, Yang and He, Jiawei and Fan, Lue and Zhang, Zhaoxiang},
  booktitle={Proceedings of the IEEE/CVF International Conference on Computer Vision},
  pages={27137--27146},
  year={2025}
}

@article{chen2024e2eadsurvey,
  title={End-to-End Autonomous Driving: Challenges and Frontiers},
  author={Chen, Li and Wu, Penghao and Chitta, Kashyap and Jaeger, Bernhard and Geiger, Andreas and Li, Hongyang},
  journal={IEEE Transactions on Pattern Analysis and Machine Intelligence},
  year={2024}
}

@inproceedings{sadat2020p3,
  title={Perceive, Predict, and Plan: Safe Motion Planning Through Interpretable Semantic Representations},
  author={Sadat, Abbas and Casas, Sergio and Ren, Mengye and Wu, Xinyu and Dhawan, Pranaab and Urtasun, Raquel},
  booktitle={European Conference on Computer Vision},
  pages={414--430},
  year={2020},
  organization={Springer}
}

@inproceedings{hu2022stp3,
  title={{ST-P3}: End-to-End Vision-Based Autonomous Driving via Spatial-Temporal Feature Learning},
  author={Hu, Shengchao and Chen, Li and Wu, Penghao and Li, Hongyang and Yan, Junchi and Tao, Dacheng},
  booktitle={European Conference on Computer Vision},
  pages={533--549},
  year={2022},
  organization={Springer}
}

@inproceedings{wu2022tcp,
  title={Trajectory-Guided Control Prediction for End-to-End Autonomous Driving: A Simple Yet Strong Baseline},
  author={Wu, Penghao and Jia, Xiaosong and Chen, Li and Yan, Junchi and Li, Hongyang and Qiao, Yu},
  booktitle={Advances in Neural Information Processing Systems},
  year={2022}
}

@article{yuan2024drama,
  title={DRAMA: An Efficient End-to-End Motion Planner for Autonomous Driving with Mamba},
  author={Yuan, Chengran and Zhang, Zhanqi and Sun, Jiawei and Sun, Shuo and Huang, Zefan and Lee, Christina Dao Wen and Li, Dongen and Han, Yuhang and Wong, Anthony and Tee, Keng Peng and others},
  journal={arXiv preprint arXiv:2408.03601},
  year={2024}
}

@inproceedings{dauner2024navsim,
  title={NAVSIM: Data-Driven Non-Reactive Autonomous Vehicle Simulation and Benchmarking},
  author={Dauner, Daniel and Hallgarten, Marcel and Li, Tianyu and Weng, Xinshuo and Huang, Zhiyu and Yang, Zetong and Li, Hongyang and Gilitschenski, Igor and Ivanovic, Boris and Pavone, Marco and Geiger, Andreas and Chitta, Kashyap},
  booktitle={Advances in Neural Information Processing Systems},
  volume={37},
  pages={28706--28719},
  year={2024}
}

@inproceedings{dauner2023pdmclosed,
  title={Parting with Misconceptions about Learning-Based Vehicle Motion Planning},
  author={Dauner, Daniel and Hallgarten, Marcel and Geiger, Andreas and Chitta, Kashyap},
  booktitle={Conference on Robot Learning},
  pages={1268--1281},
  year={2023},
  organization={PMLR}
}

@article{fan2018apollo,
  title={Baidu Apollo EM Motion Planner},
  author={Fan, Haoyang and Zhu, Fan and Liu, Changchun and Zhang, Liangliang and Zhuang, Li and Li, Dong and Zhu, Weicheng and Hu, Jiangtao and Li, Hongye and Kong, Qi},
  journal={arXiv preprint arXiv:1807.08048},
  year={2018}
}

@article{treiber2000congested,
  title={Congested Traffic States in Empirical Observations and Microscopic Simulations},
  author={Treiber, Martin and Hennecke, Ansgar and Helbing, Dirk},
  journal={Physical Review E},
  volume={62},
  number={2},
  pages={1805--1824},
  year={2000},
  publisher={APS}
}

@inproceedings{chen2021worldonrails,
  title={Learning to Drive from a World on Rails},
  author={Chen, Dian and Koltun, Vladlen and Kr{\"a}henb{\"u}hl, Philipp},
  booktitle={Proceedings of the IEEE/CVF Conference on Computer Vision and Pattern Recognition},
  pages={15590--15599},
  year={2021}
}

@inproceedings{henaff2019mppuu,
  title={Model-Predictive Policy Learning with Uncertainty Regularization for Driving in Dense Traffic},
  author={Henaff, Mikael and Canziani, Alfredo and LeCun, Yann},
  booktitle={International Conference on Learning Representations},
  year={2019}
}

@inproceedings{qiao2024marlcce,
  title={Modelling Competitive Behaviors in Autonomous Driving Under Generative World Model},
  author={Qiao, Guanren and Quan, Guorui and Qu, Rongxiao and Liu, Guiliang},
  booktitle={European Conference on Computer Vision},
  pages={19--36},
  year={2024},
  organization={Springer}
}

@article{hu2023gaia1,
  title={{GAIA-1}: A Generative World Model for Autonomous Driving},
  author={Hu, Anthony and Russell, Lloyd and Yeo, Hudson and Murez, Zak and Fedoseev, George and Kendall, Alex and Shotton, Jamie and Corrado, Gianluca},
  journal={arXiv preprint arXiv:2309.17080},
  year={2023}
}

@inproceedings{wang2023drivedreamer,
  title={{DriveDreamer}: Towards Real-World-Driven World Models for Autonomous Driving},
  author={Wang, Xiaofeng and Zhu, Zheng and Huang, Guan and Chen, Xinze and Zhu, Jiagang and Lu, Jiwen},
  booktitle={European Conference on Computer Vision},
  year={2024}
}

@inproceedings{gao2024vista,
  title={Vista: A Generalizable Driving World Model with High Fidelity and Versatile Controllability},
  author={Gao, Shenyuan and Yang, Jiazhi and Chen, Li and Chitta, Kashyap and Qiu, Yihang and Geiger, Andreas and Zhang, Jun and Li, Hongyang},
  booktitle={Advances in Neural Information Processing Systems},
  year={2024}
}

@article{zhang2024bevworld,
  title={{BEVWorld}: A Multimodal World Model for Autonomous Driving via Unified {BEV} Latent Space},
  author={Zhang, Yumeng and Gong, Shi and Xiong, Kaixin and Ye, Xiaoqing and Tan, Xiao and Wang, Fan and Huang, Jizhou and Wu, Hua and Wang, Haifeng},
  journal={arXiv preprint arXiv:2407.05679},
  year={2024}
}

@article{li2025hydraplusplus,
  title={Hydra-MDP++: Advancing End-to-End Driving via Expert-Guided Hydra-Distillation},
  author={Li, Kailin and Li, Zhenxin and Lan, Shiyi and Xie, Yuan and Zhang, Zhizhong and Liu, Jiayi and Wu, Zuxuan and Yu, Zhiding and Alvarez, Jose M.},
  journal={arXiv preprint arXiv:2503.12820},
  year={2025}
}

@inproceedings{zhang2026resworld,
  title={{ResWorld}: Temporal Residual World Model for End-to-End Autonomous Driving},
  author={Zhang, Jinqing and Fu, Zehua and Xu, Zelin and Dai, Wenying and Liu, Qingjie and Wang, Yunhong},
  booktitle={The Fourteenth International Conference on Learning Representations},
  year={2026}
}

@inproceedings{hu2018senet,
  title={Squeeze-and-Excitation Networks},
  author={Hu, Jie and Shen, Li and Albanie, Samuel and Sun, Gang and Wu, Enhua},
  booktitle={Proceedings of the IEEE/CVF Conference on Computer Vision and Pattern Recognition},
  pages={7132--7141},
  year={2018}
}

@inproceedings{zhu2021deformabledetr,
  title={Deformable DETR: Deformable Transformers for End-to-End Object Detection},
  author={Zhu, Xizhou and Su, Weijie and Lu, Lewei and Li, Bin and Wang, Xiaogang and Dai, Jifeng},
  booktitle={International Conference on Learning Representations},
  year={2021}
}

@inproceedings{zhang2023sa,
  title={{SA-BEV}: Generating Semantic-Aware Bird's-Eye-View Feature for Multi-View {3D} Object Detection},
  author={Zhang, Jinqing and Zhang, Yanan and Liu, Qingjie and Wang, Yunhong},
  booktitle={Proceedings of the IEEE/CVF International Conference on Computer Vision},
  pages={3348--3357},
  year={2023}
}

@inproceedings{zhang2025geobev,
  title={{GeoBEV}: Learning Geometric {BEV} Representation for Multi-View {3D} Object Detection},
  author={Zhang, Jinqing and Zhang, Yanan and Qi, Yunlong and Fu, Zehua and Liu, Qingjie and Wang, Yunhong},
  booktitle={Proceedings of the AAAI Conference on Artificial Intelligence},
  volume={39},
  number={9},
  pages={9960--9968},
  year={2025}
}

@inproceedings{li2022hdmapnet,
  title={{HDMapNet}: An Online HD Map Construction and Evaluation Framework},
  author={Li, Qi and Wang, Yue and Wang, Yilun and Zhao, Hang},
  booktitle={IEEE International Conference on Robotics and Automation},
  pages={4628--4634},
  year={2022}
}

@article{huang2022bevdet4d,
  title={{BEVDet4D}: Exploit Temporal Cues in Multi-Camera {3D} Object Detection},
  author={Huang, Junjie and Huang, Guan},
  journal={arXiv preprint arXiv:2203.17054},
  year={2022}
}

@inproceedings{sima2023sceneasoccupancy,
  title={Scene as Occupancy},
  author={Sima, Chonghao and Tong, Wenwen and Wang, Tai and Chen, Li and Wu, Silei and Deng, Hanming and Gu, Yi and Lu, Lewei and Luo, Ping and Lin, Dahua and Li, Hongyang},
  booktitle={Proceedings of the IEEE/CVF International Conference on Computer Vision},
  pages={8406--8415},
  year={2023}
}

@article{yu2023flashocc,
  title={{FlashOCC}: Fast and Memory-Efficient Occupancy Prediction via Channel-to-Height Plugin},
  author={Yu, Zichen and Shu, Changyong and Deng, Jiajun and Lu, Kangjie and Liu, Zongdai and Yu, Jiangyong and Yang, Dawei and Li, Hui and Chen, Yan},
  journal={arXiv preprint arXiv:2311.12058},
  year={2023}
}

@inproceedings{yuan2024streammapnet,
  title={StreamMapNet: Streaming Mapping Network for Vectorized Online HD Map Construction},
  author={Yuan, Tianyuan and Liu, Yicheng and Wang, Yue and Wang, Yilun and Zhao, Hang},
  booktitle={Proceedings of the IEEE/CVF Winter Conference on Applications of Computer Vision},
  pages={7356--7365},
  year={2024}
}

@inproceedings{he2016deep,
  title={Deep Residual Learning for Image Recognition},
  author={He, Kaiming and Zhang, Xiangyu and Ren, Shaoqing and Sun, Jian},
  booktitle={Proceedings of the IEEE Conference on Computer Vision and Pattern Recognition},
  pages={770--778},
  year={2016}
}

@article{vaswani2017attention,
  title={Attention Is All You Need},
  author={Vaswani, Ashish and Shazeer, Noam and Parmar, Niki and Uszkoreit, Jakob and Jones, Llion and Gomez, Aidan N and Kaiser, {\L}ukasz and Polosukhin, Illia},
  journal={Advances in Neural Information Processing Systems},
  volume={30},
  year={2017}
}

@inproceedings{jia2024bench,
  title={{Bench2Drive}: Towards Multi-Ability Benchmarking of Closed-Loop End-To-End Autonomous Driving},
  author={Jia, Xiaosong and Yang, Zhenjie and Li, Qifeng and Zhang, Zhiyuan and Yan, Junchi},
  booktitle={NeurIPS 2024 Datasets and Benchmarks Track},
  year={2024}
}

@article{zhai2023rethinking,
  title={Rethinking the Open-Loop Evaluation of End-to-End Autonomous Driving in {nuScenes}},
  author={Zhai, Jiang-Tian and Feng, Ze and Du, Jinhao and Mao, Yongqiang and Liu, Jiang-Jiang and Tan, Zichang and Zhang, Yifu and Ye, Xiaoqing and Wang, Jingdong},
  journal={arXiv preprint arXiv:2305.10430},
  year={2023}
}

\appendix
\newpage
\section{Further Ablation Study}

\paragraph{Effect of the number of supervised future timestamps.}
We further study how many future timestamps should be supervised during training to effectively shape the shared memory. As shown in Table~\ref{tab:ablation-num-future-sup}, planning performance improves consistently as more future timestamps are supervised. For intermediate settings, timestamps are randomly sampled from both the earlier and later halves of the horizon, which encourages temporally balanced predictive shaping. This result suggests that future supervision is essential for injecting predictive temporal structure into the shared memory.

\begin{table}[H]
\centering
\small
\setlength{\tabcolsep}{4.5pt}
\caption{Ablation study on the number of supervised future timestamps. ``\#Future Sup.'' denotes the number of future timestamps used for semantic supervision during training. ``0'' corresponds to the setting with world model but without future supervision. The best results are shown in \textbf{bold}.}
\label{tab:ablation-num-future-sup}
\begin{tabular}{c c c c c c c}
\toprule
\#Future Sup. & NC$\uparrow$ & DAC$\uparrow$ & TTC$\uparrow$ & Comf.$\uparrow$ & EP$\uparrow$ & PDMS$\uparrow$ \\
\midrule
0 & 98.5 & 96.5 & 95.1 & 99.8 & 82.3 & 88.4 \\
2 & 98.8 & 97.0 & 95.7 & \textbf{100.0} & 83.7 & 89.5 \\
4 & 99.0 & 97.3 & 95.9 & 99.9 & 84.0 & 90.0 \\
6 & 99.0 & 97.4 & \textbf{96.2} & 99.9 & 84.3 & 90.3 \\
8 & \textbf{99.0} & \textbf{97.8} & 96.0 & \textbf{100.0} & \textbf{84.6} & \textbf{90.5} \\
\bottomrule
\end{tabular}
\end{table}

When no future timestamp is supervised, the world model branch is optimized only indirectly through the final planning objective, which is insufficient to reliably expose future-relevant information for downstream trajectory evaluation. Once future supervision is introduced, performance improves steadily as more timestamps are covered. We attribute this trend to the fact that a denser set of supervised future readouts provides stronger temporal constraints on the shared memory, encouraging it to encode scene evolution over a broader horizon.

\section{Additional Visualization Results}

\subsection{Readout Semantic Maps During Training}

\begin{figure*}[t]
    \centering
    \setlength{\tabcolsep}{2pt}
    \renewcommand{\arraystretch}{1.05}

    \begin{tabular}{c c c c c}
        & \footnotesize $T=1\,\mathrm{s}$ & \footnotesize $T=2\,\mathrm{s}$ & \footnotesize $T=3\,\mathrm{s}$ & \footnotesize $T=4\,\mathrm{s}$ \\

        \rotatebox{90}{\footnotesize Prediction}
        & \includegraphics[width=0.23\linewidth]{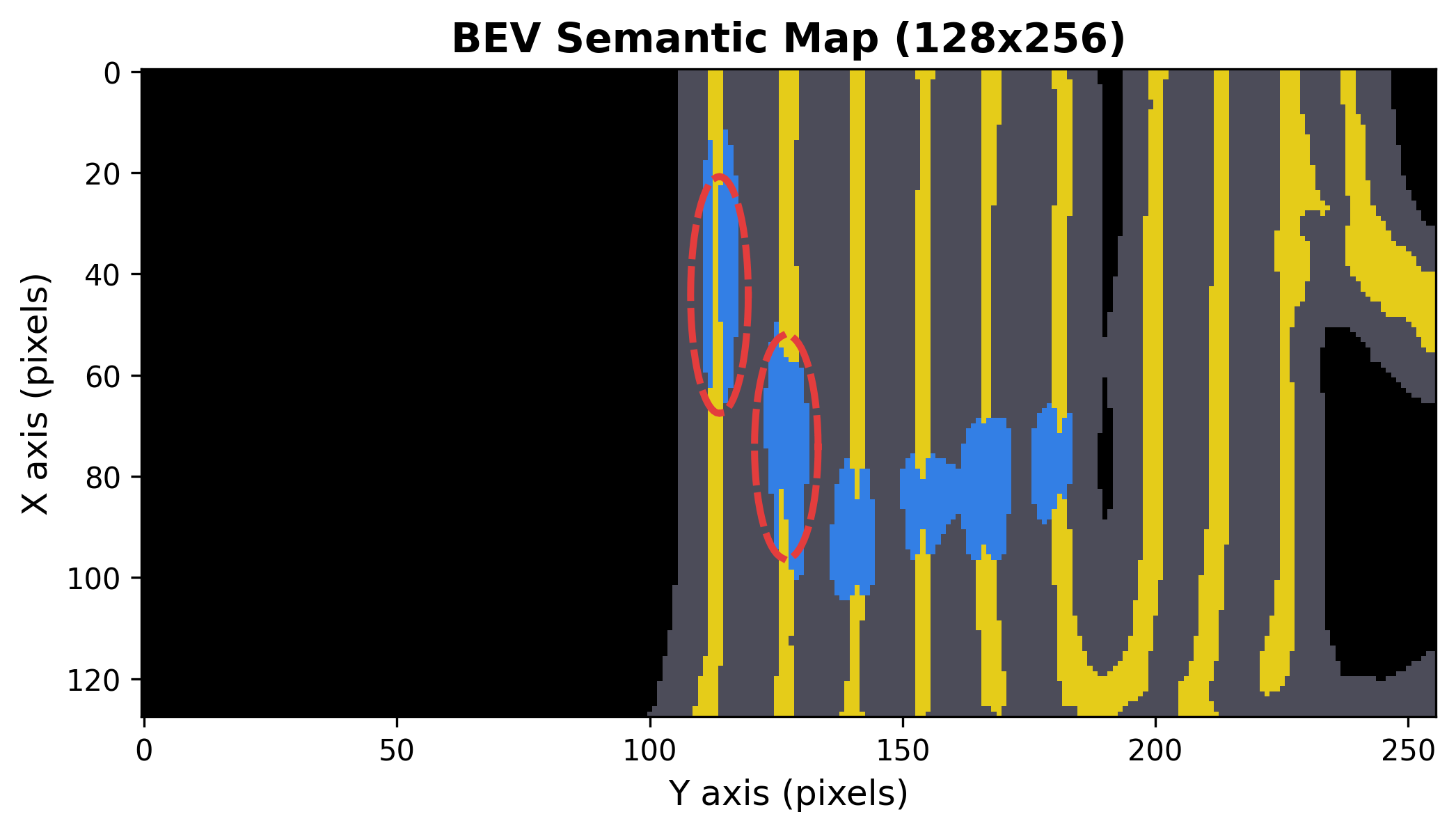}
        & \includegraphics[width=0.23\linewidth]{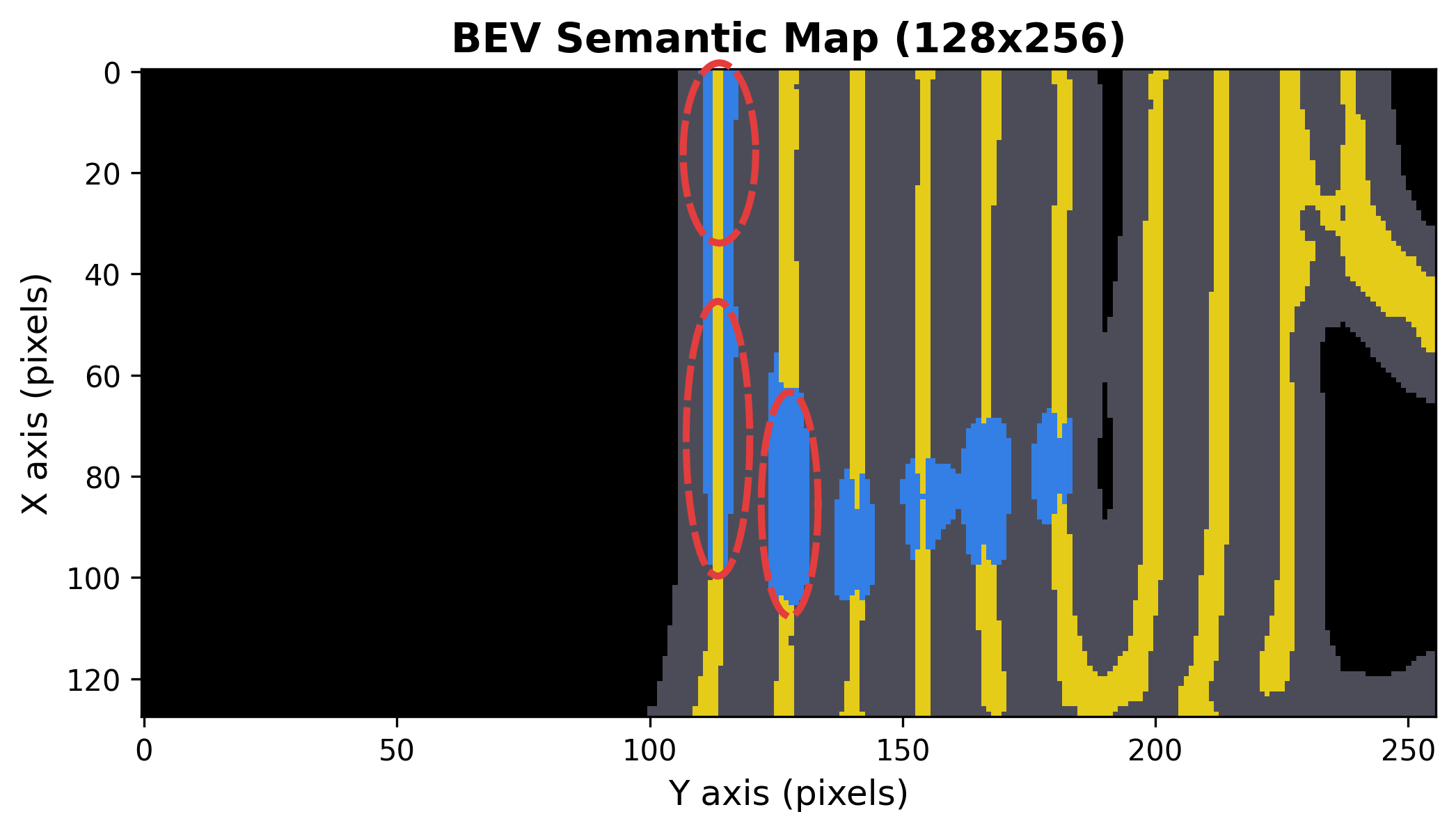}
        & \includegraphics[width=0.23\linewidth]{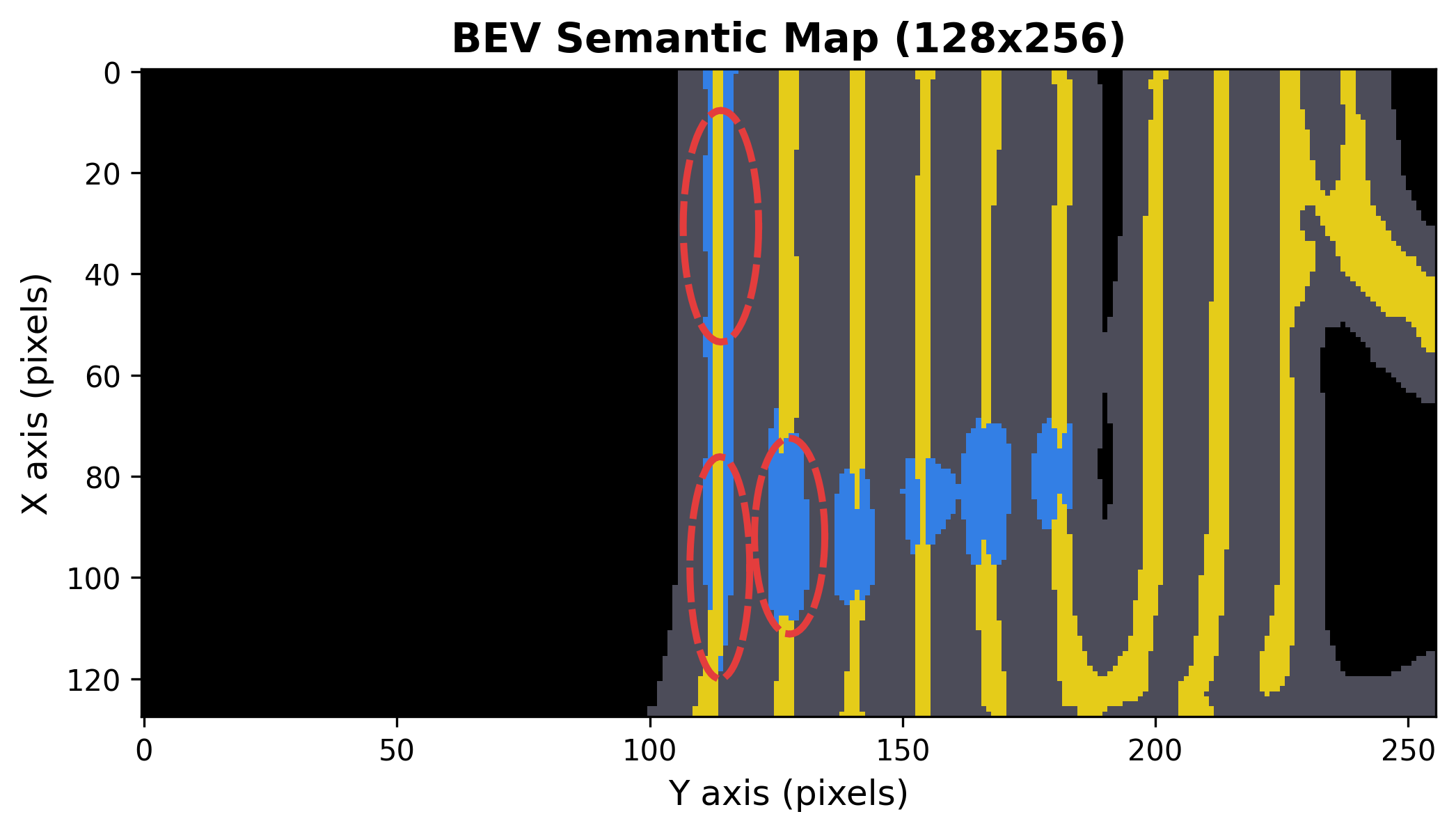}
        & \includegraphics[width=0.23\linewidth]{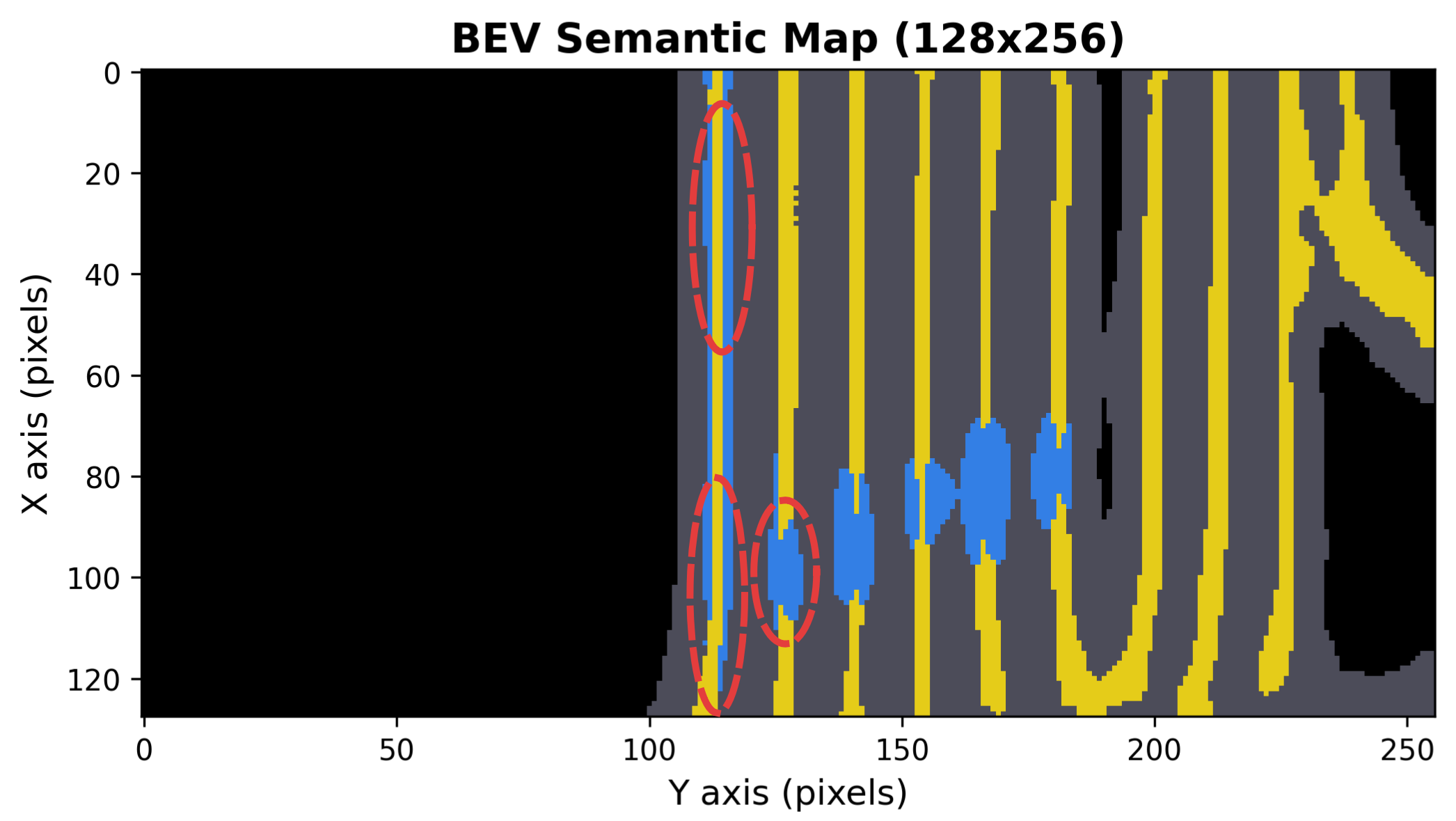} \\

        \rotatebox{90}{\footnotesize Ground Truth}
        & \includegraphics[width=0.23\linewidth]{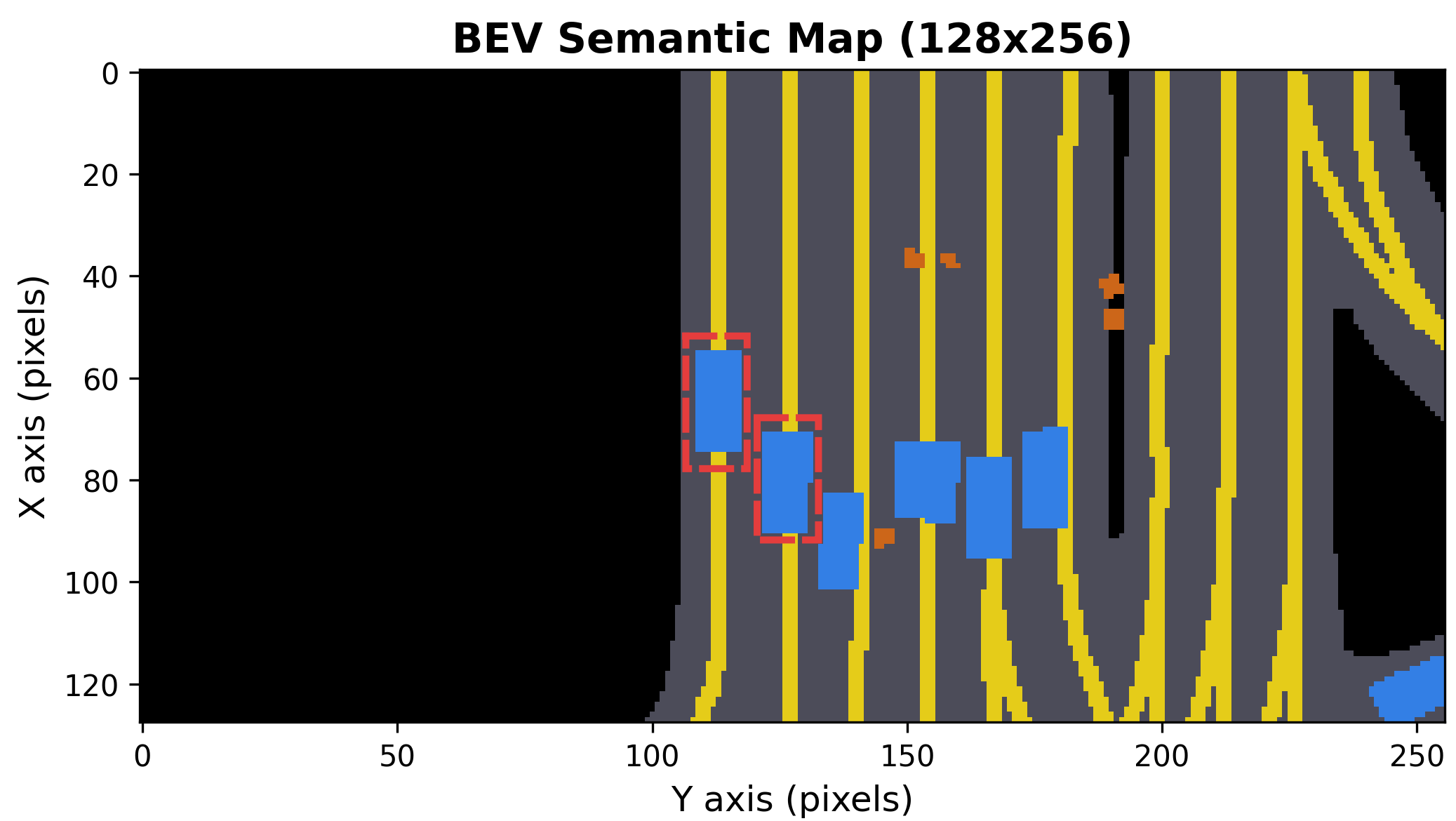}
        & \includegraphics[width=0.23\linewidth]{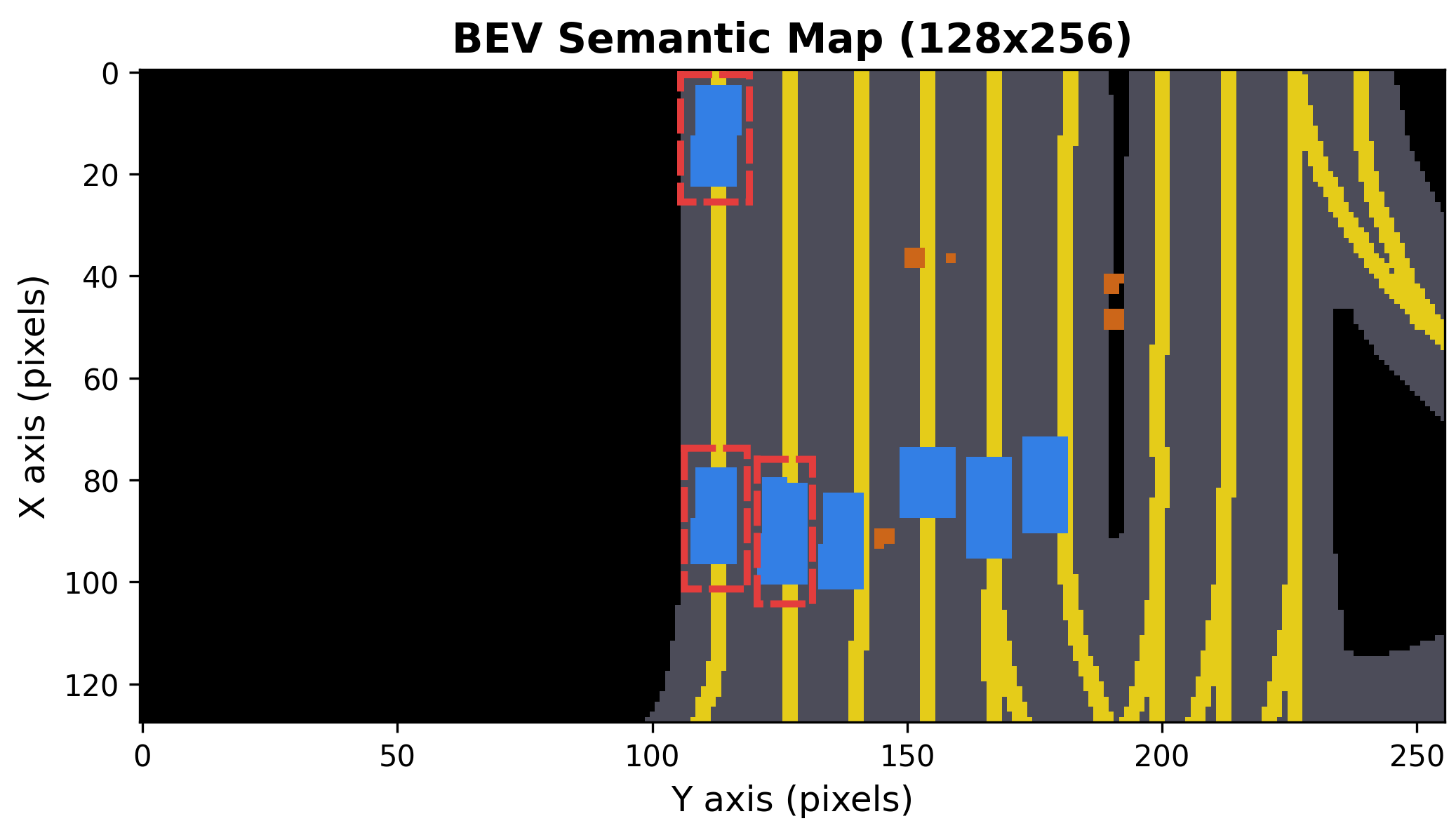}
        & \includegraphics[width=0.23\linewidth]{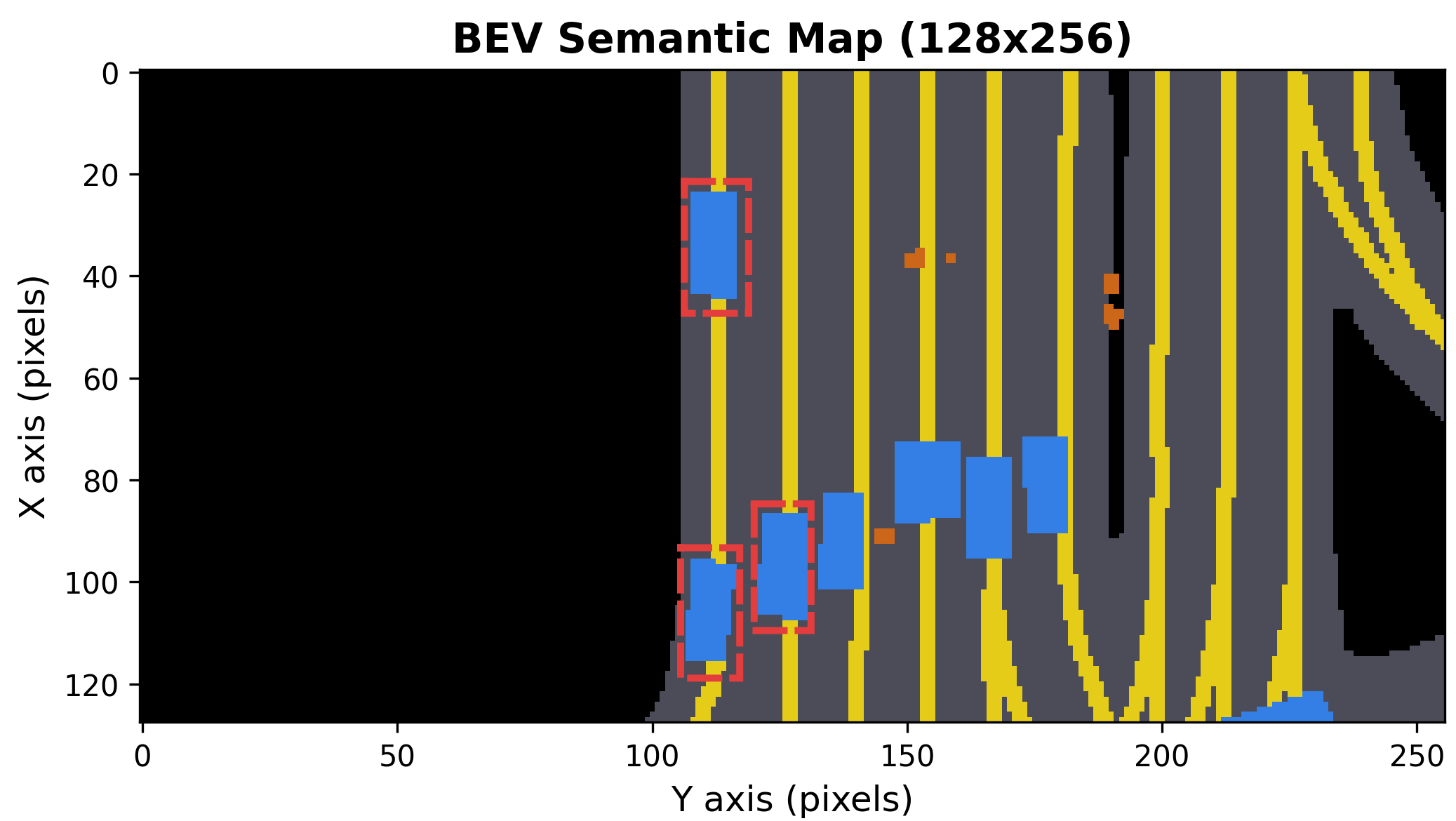}
        & \includegraphics[width=0.23\linewidth]{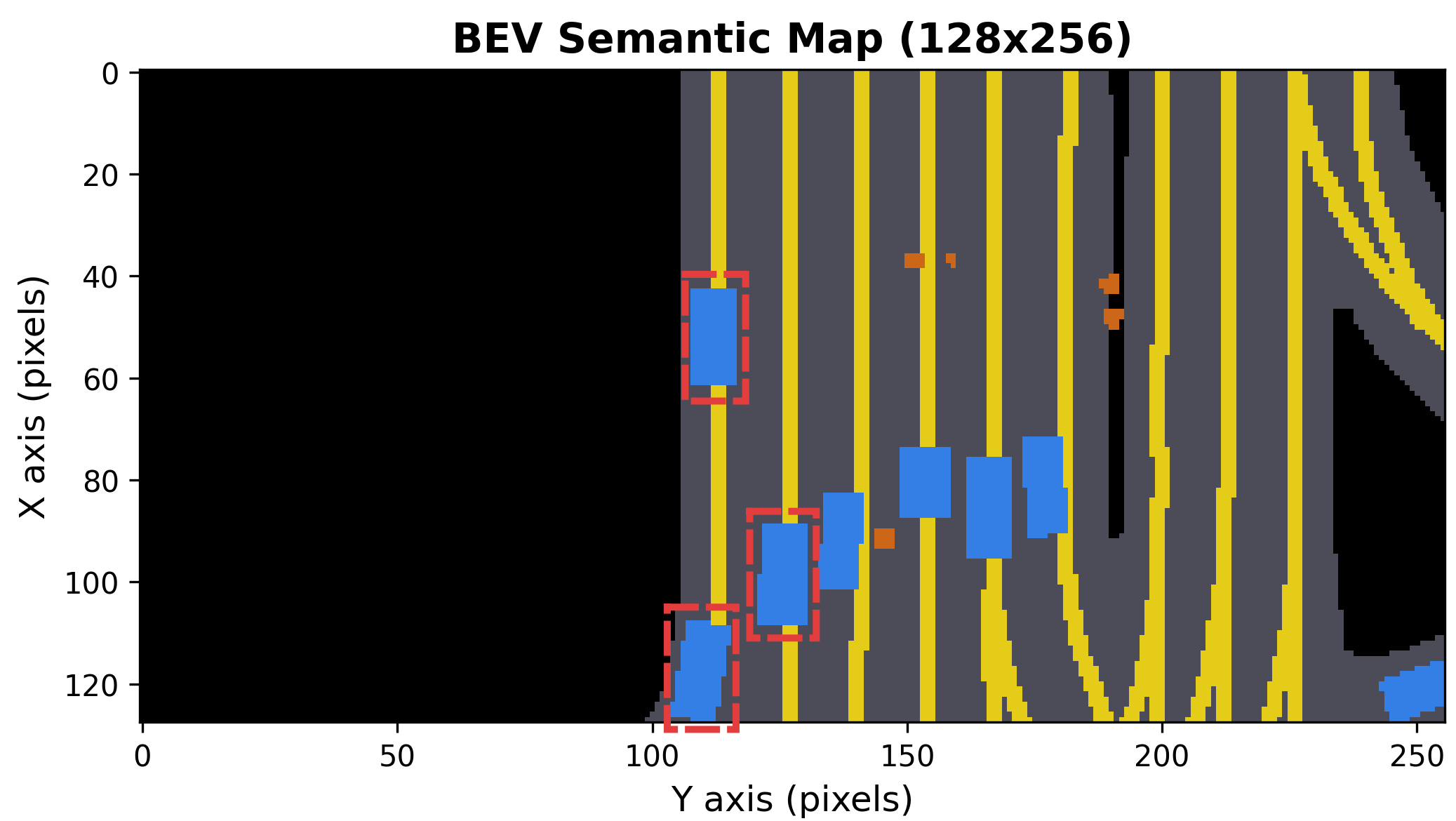} \\
    \end{tabular}

    \vspace{4pt}

    {\footnotesize
    \semlegenditem{black}{Background}\hspace{1.0em}
    \semlegenditem{semroad}{Road (Drivable)}\hspace{1.0em}
    \semlegenditem{semwalk}{Walkways}\hspace{1.0em}
    \semlegenditem{semcenter}{Centerline}\hspace{1.0em}
    \semlegenditem{semstatic}{Static Objects}\hspace{1.0em}
    \semlegenditem{semvehicle}{Vehicles}
    }

    \caption{Visualization of the readout semantic maps during training. The top row shows the semantic maps predicted from the time-conditioned readout features, and the bottom row shows the corresponding ground-truth semantic maps. From left to right, the columns correspond to four future timestamps at $T=1\,\mathrm{s}$, $2\,\mathrm{s}$, $3\,\mathrm{s}$, and $4\,\mathrm{s}$. Red dashed boxes highlight representative dynamic regions to facilitate comparison of their temporal evolution.}
    \label{fig:appendix-semantic-readout}
\end{figure*}

To further examine whether the training-time temporal readout indeed shapes the shared memory into a future-aware representation, we visualize in Fig.~\ref{fig:appendix-semantic-readout} the semantic maps decoded from the readout features at four future timestamps. To make the temporal differences easier to inspect, representative dynamic objects are highlighted by red dashed boxes in both the predicted maps and the corresponding ground truth maps. Since the semantic decoder is shared across all timestamps, the differences across columns are induced solely by the time-conditioned readout rather than by timestamp-specific decoding heads.

As shown in Fig.~\ref{fig:appendix-semantic-readout}, the predicted maps consistently preserve the major scene semantics, including the drivable region, centerline structure, and the coarse layout of surrounding vehicles. More importantly, the highlighted regions show that the readout features capture temporally evolving vehicle configurations across future timestamps, rather than simply reproducing the current scene. Although minor deviations in shape and occupancy remain, the overall temporal evolution is largely consistent with the ground truth.

These observations provide qualitative evidence that temporal readout with future semantic supervision injects predictive temporal structure into the shared memory. In other words, the readout branch learns semantically meaningful future features during training, which helps shape a planning-oriented memory for subsequent trajectory evaluation.

\subsection{Temporal Structure of Readout Gates}

To further understand how the temporal readout modulates the shared memory across timestamps, we analyze the learned readout gates from two complementary perspectives. We first examine channel-wise temporal modulation, which reveals how individual channels are reweighted when the readout moves from the current timestamp to future timestamps. We then study the global statistics of the gate distribution, which summarize how the overall selectivity of temporal readout evolves over the prediction horizon.

\subsubsection{Channel-wise Temporal Modulation}

\begin{figure*}[t]
    \centering
    \setlength{\tabcolsep}{2pt}
    \begin{subfigure}[t]{0.52\textwidth}
        \centering
        \includegraphics[width=\linewidth]{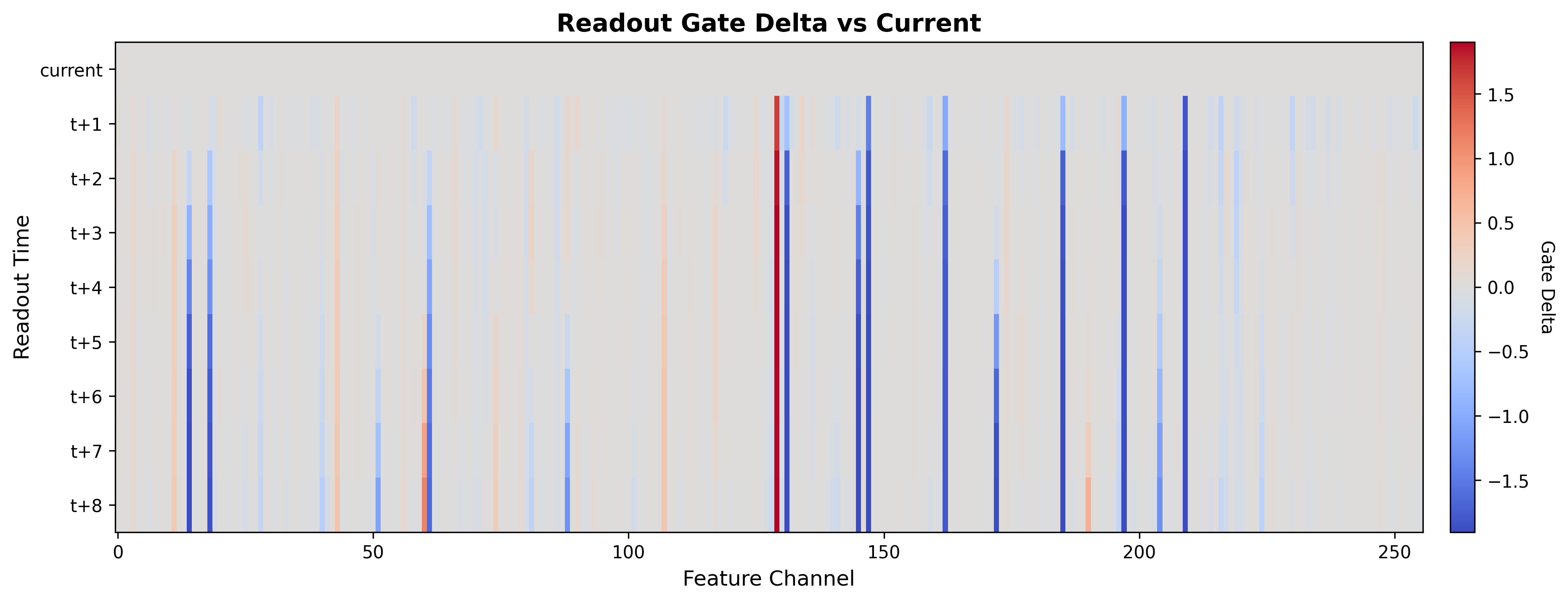}
        \caption{Channel-wise gate differences relative to the current timestamp.}
        \label{fig:appendix-gate-delta}
    \end{subfigure}
    \hfill
    \begin{subfigure}[t]{0.45\textwidth}
        \centering
        \includegraphics[width=\linewidth]{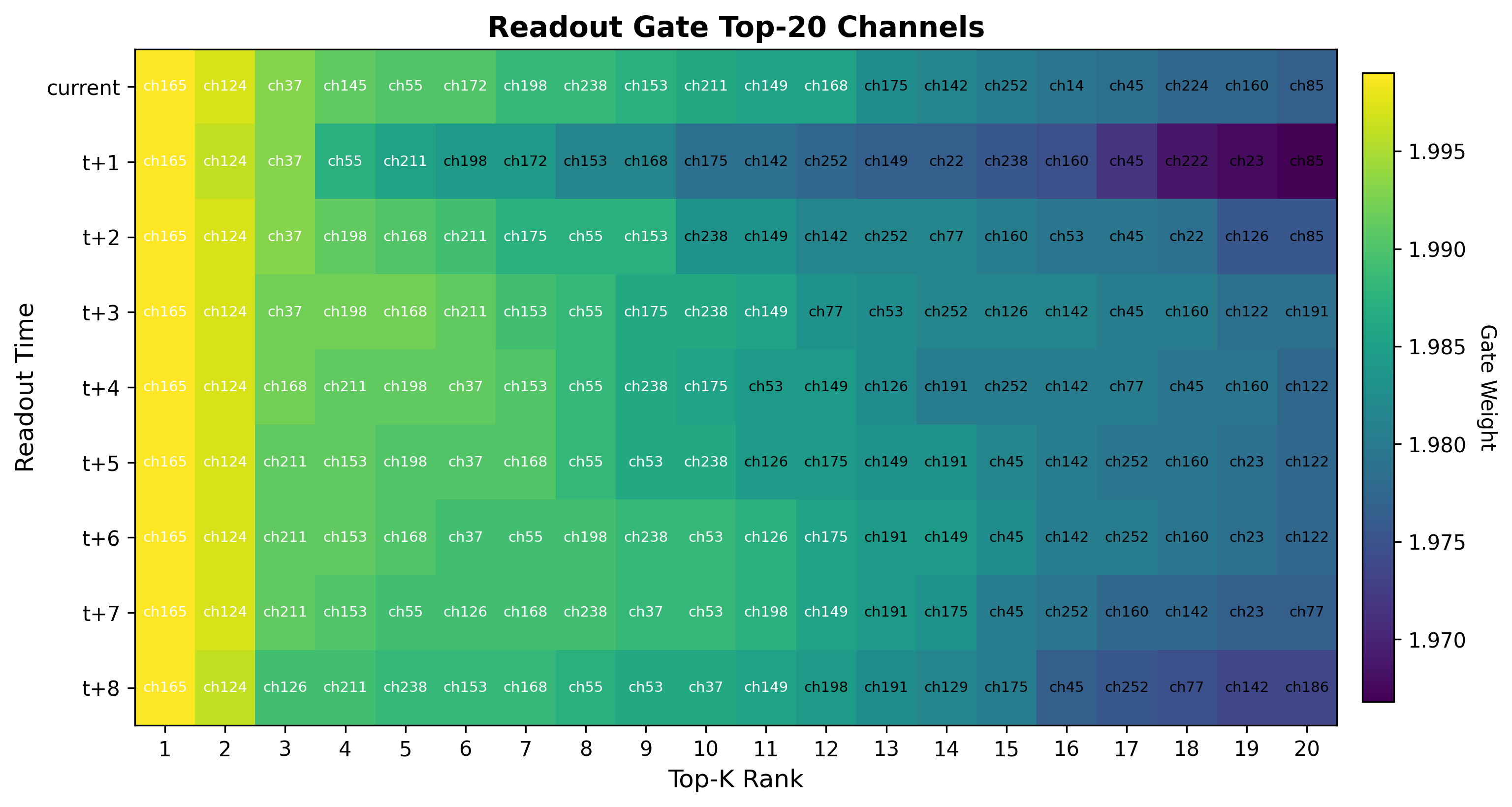}
        \caption{Top-20 channels with the largest gate values at each timestamp.}
        \label{fig:appendix-gate-topk}
    \end{subfigure}
    \caption{Channel-wise analysis of temporal readout gates. Left: gate differences between each future timestamp and the current timestamp. Right: Top-20 channels ranked by gate value at each timestamp.}
    \label{fig:appendix-gate-channelwise}
\end{figure*}

We first investigate the channel-level behavior of temporal readout in Fig.~\ref{fig:appendix-gate-channelwise}. The gate-delta heatmap visualizes, for each future timestamp, the difference between its gate values and those at the current timestamp. A clear pattern can be observed: most channels remain close to the current readout, while only a relatively small subset exhibits noticeable positive or negative deviations. This suggests that temporal readout does not construct future representations independently for each timestamp. Instead, it performs structured modulation on top of a shared memory, preserving a common channel basis while selectively adjusting a subset of channels according to the queried timestamp.

The Top-20 visualization provides a complementary view of this behavior. Several highly weighted channels remain consistently active across timestamps, indicating that part of the channel basis is shared by all temporal readouts. Meanwhile, the lower-ranked channels change gradually as the timestamp moves farther into the future, showing that temporal conditioning mainly acts by refining the supporting channels around a stable core set. Taken together, these results indicate that the temporal readout follows a ``shared basis + timestamp-specific modulation'' pattern, which is consistent with the design goal of RFWorld: the shared memory stores reusable scene structure, while the readout gate exposes timestamp-relevant predictive information through lightweight channel reweighting.

\subsubsection{Distributional Statistics Across Timestamps}

\begin{figure*}[t]
    \centering
    \setlength{\tabcolsep}{2pt}
    \begin{subfigure}[t]{0.485\textwidth}
        \centering
        \includegraphics[width=\linewidth]{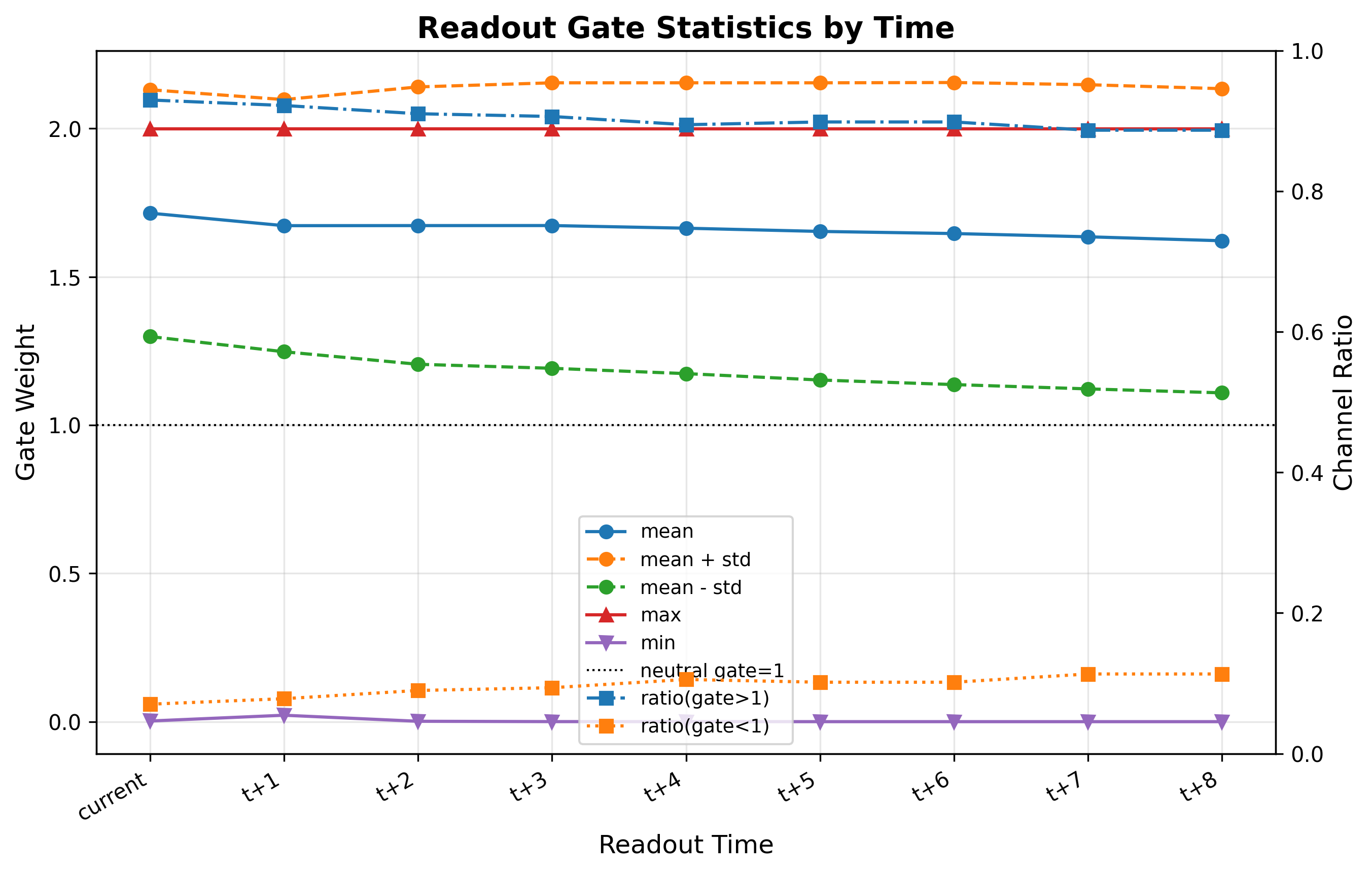}
        \caption{Summary statistics of gate values across timestamps.}
        \label{fig:appendix-gate-stats}
    \end{subfigure}
    \hfill
    \begin{subfigure}[t]{0.485\textwidth}
        \centering
        \includegraphics[width=\linewidth]{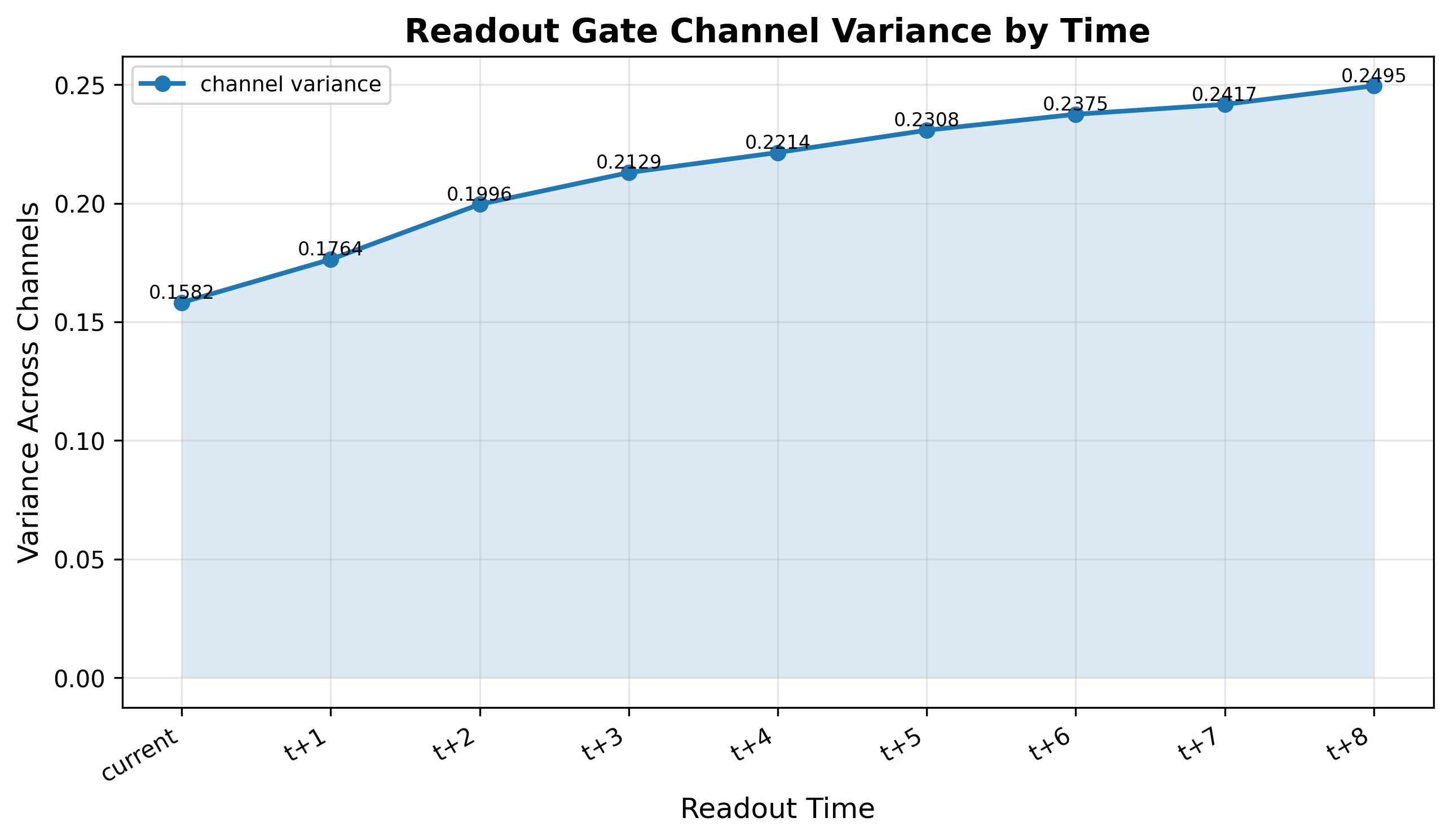}
        \caption{Variance of gate values across channels at each timestamp.}
        \label{fig:appendix-gate-variance}
    \end{subfigure}
    \caption{Distributional analysis of temporal readout gates. Left: aggregated statistics of gate values at each timestamp. Right: channel-wise variance of gate values at each timestamp.}
    \label{fig:appendix-gate-statistics}
\end{figure*}

We next summarize the temporal readout behavior at the distribution level in Fig.~\ref{fig:appendix-gate-statistics}. The gate statistics show that the mean gate value stays above the neutral value across all timestamps, and that the proportion of channels with gate values larger than the neutral value consistently dominates the proportion below it. This indicates that the temporal readout mainly operates by enhancing informative channels rather than suppressing the majority of channels. At the same time, as the readout timestamp moves farther into the future, the mean gate value decreases slightly and the fraction of down-weighted channels grows modestly, suggesting that longer-horizon readout becomes progressively more selective.

This tendency is further corroborated by the variance plot. The variance of gate values across channels increases steadily with the timestamp, indicating that the channel weights become more dispersed for farther-future readout. In other words, the readout gate applies increasingly differentiated reweighting over channels as the timestamp advances. This observation complements the channel-wise analysis above: near-future readout remains closer to the current feature basis, whereas farther-future readout requires stronger channel discrimination to extract more predictive temporal information from the shared memory.

Overall, the two levels of visualization form a coherent picture. Temporal readout in RFWorld is neither static nor arbitrary. It preserves a stable shared channel basis across timestamps, while gradually increasing channel-wise selectivity as the queried timestamp moves farther into the future. This behavior supports the main-text claim that future supervision shapes a shared memory with predictive temporal structure, and that such structure can be exposed efficiently through sparse timestamp-conditioned readout rather than explicit future-state rollout.

\end{document}